\pdfoutput=1

\documentclass[11pt]{article}

\usepackage[final]{acl}

\usepackage{times}
\usepackage{todonotes}
\usepackage{latexsym}
\usepackage{amsmath,amssymb,amsfonts}%
\usepackage[T1]{fontenc}

\usepackage[utf8]{inputenc}
\usepackage{bm}
\usepackage{tcolorbox}

\usepackage{microtype}
\usepackage{array}
\usepackage{multirow}%
\usepackage{xcolor}
\usepackage{colortbl} 
\usepackage{inconsolata}
\usepackage{booktabs}%
\usepackage{subcaption}

\usepackage{graphicx}
\usepackage{algorithm}
\usepackage{algpseudocode}
\usepackage{xspace}
\newcommand{\methodit}{{HD-NDEs}\xspace}

\newcommand{\cde}{Neural CDEs\xspace}
\newcommand{\ode}{Neural ODEs\xspace}
\newcommand{\sde}{Neural SDEs\xspace}

\title{\methodit: Neural Differential Equations \\ for Hallucination Detection in LLMs}
\definecolor{lightgray}{gray}{0.9}
\definecolor{lightred}{rgb}{0.98, 0.81, 0.82}
\definecolor{lightgreen}{rgb}{0.88, 1.0, 0.78}
\definecolor{lightyellow}{rgb}{1.0, 1.0, 0.68}

\author{
Qing Li\textsuperscript{1}\thanks{Equal contribution.} \quad
Jiahui Geng \textsuperscript{1}\footnotemark[1]\thanks{Corresponding author: \texttt{jiahui.geng@mbzuai.ac.ae}} \quad
Zongxiong Chen\textsuperscript{2} \quad
Derui Zhu\textsuperscript{3} \quad
\textbf{Yuxia Wang}\textsuperscript{1} \\ 
\textbf{Congbo Ma}\textsuperscript{1, 4} \quad
\textbf{Chenyang Lyu}\textsuperscript{5} \quad
\textbf{Fakhri Karray}\textsuperscript{1} \\
\textsuperscript{1}Mohamed bin Zayed University of Artificial Intelligence (MBZUAI) \\
\textsuperscript{2}Fraunhofer Institute for Open Communication Systems (FOKUS) \\
\textsuperscript{3}Technical University of Munich \quad
\textsuperscript{4}New York University Abu Dhabi \\
\textsuperscript{5}Alibaba International Digital Commerce 
}

\begin{document}
\maketitle
\begin{abstract}

In recent years, large language models (LLMs) have made remarkable advancements, yet hallucination, where models produce inaccurate or non-factual statements, remains a significant challenge for real-world deployment. Although current classification-based methods, such as SAPLMA, are highly efficient in mitigating hallucinations, they struggle when non-factual information arises in the early or mid-sequence of outputs, reducing their reliability. To address these issues, we propose \textbf{H}allucination \textbf{D}etection-\textbf{N}eural \textbf{D}ifferential \textbf{E}quations (\textbf{\methodit}), a novel method that systematically assesses the truthfulness of statements by capturing the full dynamics of LLMs within their latent space. Our approaches apply neural differential equations (Neural DEs) to model the dynamic system in the latent space of LLMs. Then, the sequence in the latent space is mapped to the classification space for truth assessment. The extensive experiments across five datasets and six widely used LLMs demonstrate the effectiveness of \methodit, especially, achieving over 14\% improvement in AUC-ROC on the True-False dataset compared to state-of-the-art techniques.


\end{abstract}

\section{Introduction}

Hallucination has been widely recognized as a significant challenge in large language models (LLMs), as highlighted in various studies applications~\cite{li2023halueval, MinKLLYKIZH23, geng2023survey}. Efforts to mitigate this issue have led to the development of hallucination detection techniques, which are broadly categorized into evidence-based and evidence-free approaches. Evidence-based methods~\cite{wang2023factcheck, safe} generally involve retrieving relevant information from external sources to verify whether inconsistencies exist between the generated content and the retrieved evidence. Nevertheless, this retrieval and verification process is computationally intensive and time-consuming, making it impractical for high-throughput applications in routine use. In contrast, evidence-free methods~\cite{chen2024inside, duan2023shifting, geng2023survey} primarily utilize the inherent characteristics of LLMs and semantic features to identify potential hallucinations. These methods can be further categorized into logit-based, consistency-based, classification-based approaches, and so on. For instance, logit-based methods~\cite{huang2023look} estimate the overall uncertainty of a sentence by analyzing logit-based uncertainty at the token level. Alternatively, consistency-based methods~\cite{manakul2023selfcheckgpt} assess the consistency of model outputs, based on the premise that hallucination tends to increase variability in the generated responses.


\begin{figure}[t] 
\centering
\includegraphics[width=0.95\linewidth]{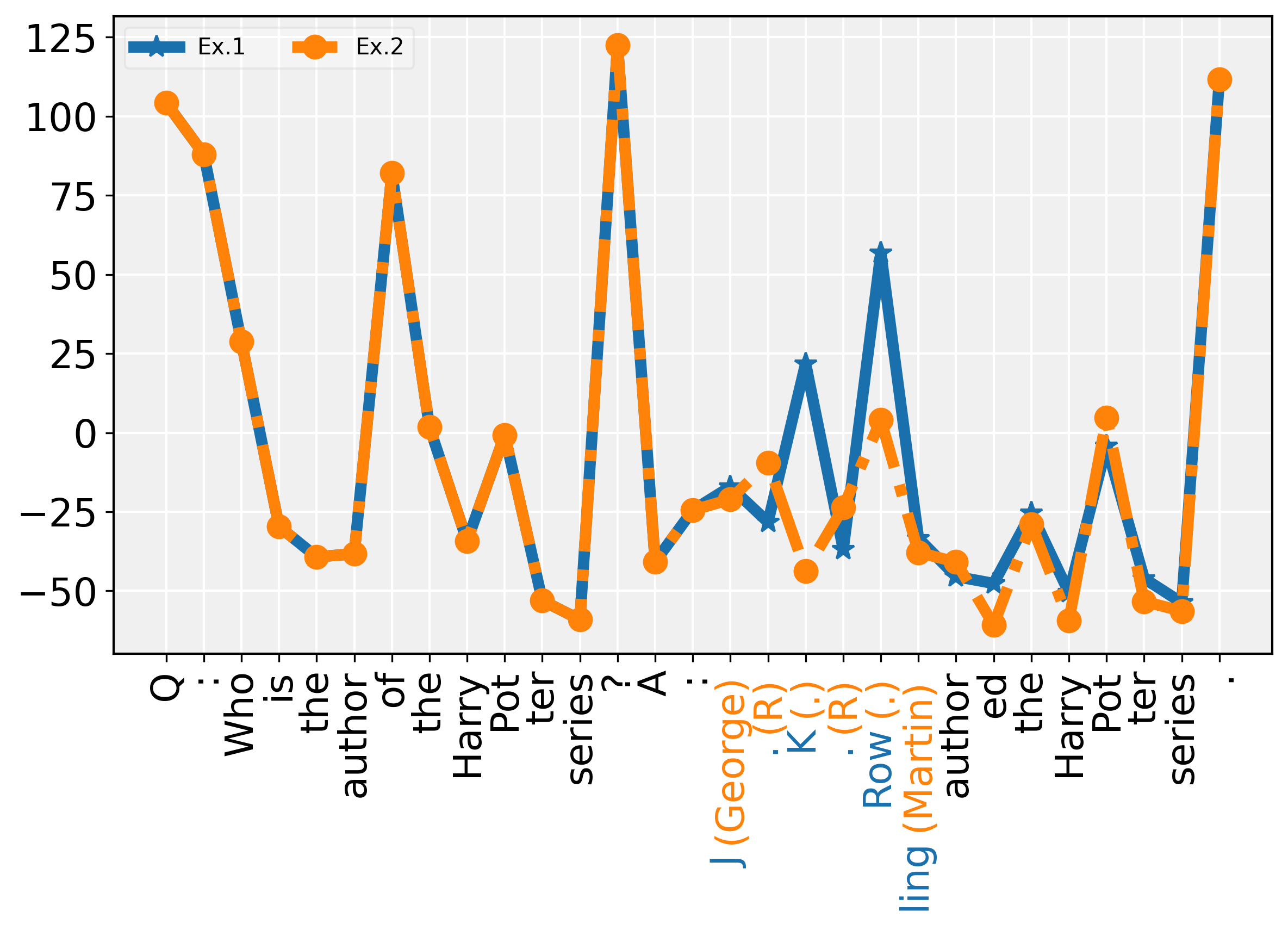} 
\caption{1D PCA projection of hidden layer embeddings for each token in Ex.1 and Ex.2. Both examples use the same question with different answers. In the hidden state space, the embeddings of the earlier tokens are identical until the tokens begin to differ. The final few tokens, being the same, result in minimal differences in the hidden state activations.}
\label{fig:ex}
\end{figure}

Furthermore, classification-based methods have proposed using a model’s internal states to probe its confidence in factual vs. non-factual sentences. Specifically, a simple feed-forward neural network classifier can be trained on the activations of the final token's last-layer hidden state to predict the reliability of the model's output. This type of method has demonstrated significant effectiveness across various model architectures, as validated by multiple  studies~\cite{azaria-mitchell-2023-internal,li2024reference,kossen2024semantic,su-etal-2024-unsupervised}. However, the classification-based method is still in its early stages and remains inadequate for handling cases where the final token of a statement fails to capture the reliability of the entire sequence. It often struggles when non-factual tokens are located at the beginning or middle of the sequence, as mentioned in ~\citet{levinstein2024still}. 
We employ principal component analysis (PCA,~\citealt{abdi2010principal}) to further investigate such failure cases. As shown in Figure~\ref{fig:ex}, we reduce the dimensionality of each token’s activations to a single dimension for clearer interpretation. Ex.1 illustrates a question with the correct answer, while Ex.2 presents the same question with an incorrect answer. Notably, the reduced hidden information of the last tokens in both examples appears nearly identical, despite differences in the middle of the sequences. This suggests that we need to effectively leverage hidden state information across the entire sequence, rather than only the last token, to accurately assess the truthfulness.

\begin{figure*}[t!]
    \centering
    \includegraphics[width=0.90\linewidth]{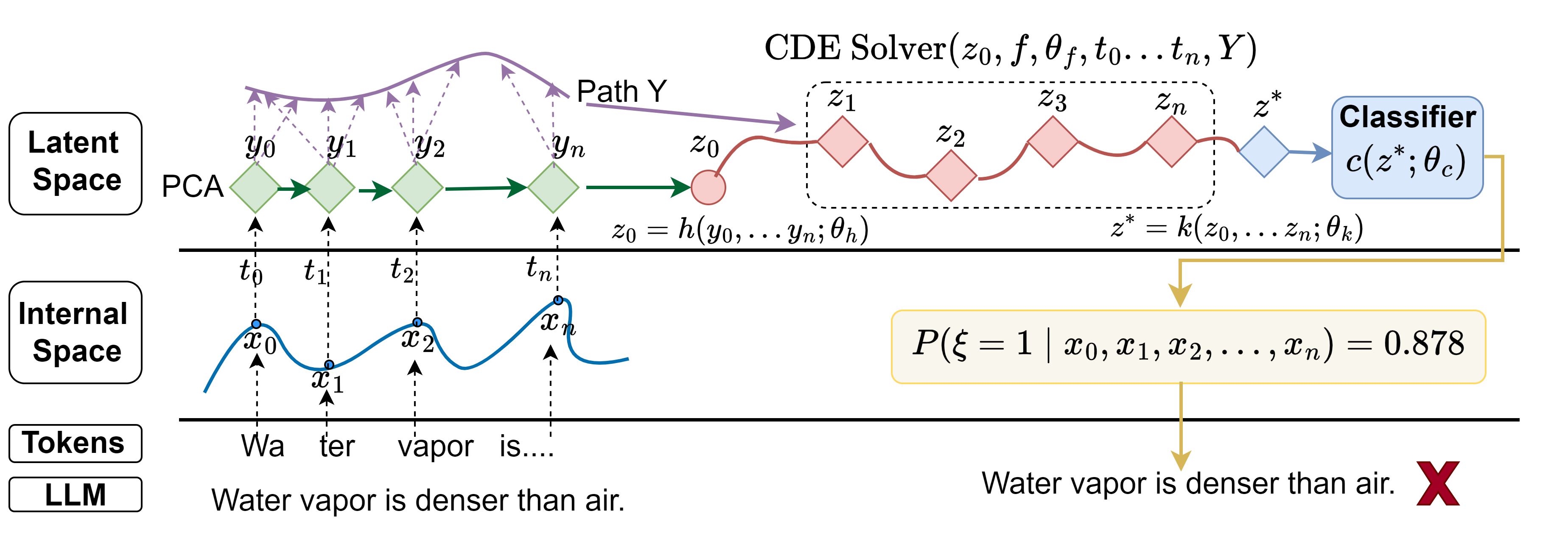}
    \caption{Computation graph of \methodit detecting hallucination via Neural CDEs. The statement is processed by LLMs, from which we extract embedding information of each token in the internal space to construct the trajectory $\bm{x}=(x_0, x_1, x_2, ..., x_n)$, with corresponding time points $(t_0, t_1,...,t_n)$. The PCA processes the trace and generates states $\bm{y}=(y_0, y_1, y_2, ..., y_n)$. These states are used to parameterize a latent space representation $z_0$ and extract control path $Y$. The CDE Solver predicts future latent states, forming $\bm{z}=(z_0, z_1, z_2, ..., z_n)$. From these latent states, $z^{*}$ is derived using the function $k$ parameterized by $\theta_{k}$. $z^{*}$ is then passed through a simple classifier to produce the final incorrect factuality score $P(\xi=1|\bm{x})$.}
    \label{fig:all:methods}
\end{figure*}

Neural differential equations (Neural DEs) have demonstrated strong effectiveness in modeling dynamic systems. Empirical studies by \citet{DBLP:journals/corr/abs-2402-14989,DBLP:conf/ijcai/LiangOYWTZ21}
demonstrate that Neural DEs outperform traditional approaches such as RNNs, LSTMs, and GRUs in capturing dynamic processes. From a theoretical standpoint, \citet{lu2019understanding,li-etal-2022-ode,baier2020n} reveal that transformers can be mathematically interpreted as numerical solvers for differential equations. The advances of Neural DEs offer a promising solution by modeling hidden state transformations as continuous trajectories, providing a more accurate representation of information flow through LLMs~\cite{kidger2022neural}. Based on effective in tasks such as time-series forecasting, classification, and outlier detection~\cite{choi2022graph,jhin2024attentive}, Neural DEs are well-suited for addressing hallucination detections in LLMs, where subtle errors can result in factual inaccuracies in generated sequences. Motivated by these strengths, our work introduces a novel, supervised method, called \methodit, marking the first application of Neural DEs in hallucination detection. As shown in Figure~\ref{fig:all:methods}, the method explicitly models the trajectory of intermediate states in the latent space using Neural DEs. Unlike previous methods that focus on individual token representations, our approach leverages temporal information in state dynamics. We conduct an extensive study on five challenging hallucination datasets, evaluating our method and state-of-the-art approaches using six widely adopted LLMs. The results demonstrate the effectiveness of our approach. Our contributions are summarized as follows:


\begin{itemize}
    \item We introduce \methodit, the first method to apply Neural DEs, including neural ordinary differential equations (Neural ODEs,~\citealt{chen2018neural}), neural controlled differential equations (Neural CDEs,~\citealt{DBLP:conf/nips/KidgerMFL20}), and neural stochastic differential equations (Neural SDEs,~\citealt{DBLP:journals/corr/abs-2402-14989}), for detecting hallucinations in LLMs. By modeling the token generation process as continuous trajectories in latent space, \methodit provides a more accurate and dynamic approach to detecting hallucinations.
    \item We evaluate \methodit on five diverse and complex hallucination datasets and compare their performance with baseline methods across six widely used LLMs. Our results demonstrate that \methodit outperforms existing approaches with a 14\% improvement in True-False Dataset.
\end{itemize}

\section{Related Work}
\paragraph{Hallucination Detection.} Hallucinations in LLMs pose significant challenges for their deployment~\cite{zhang2023siren,li2023halueval}. The generation of inaccurate information can result in customer attrition or legal risks, rendering the decision-making process unreliable. Detecting hallucinations has garnered increasing attention, and this detection is typically performed in one of the following ways: conducting a conventional retrieval task \cite{MinKLLYKIZH23,wang2023factcheck}, which requires external knowledge; converting the logits output into an uncertainty estimate for the sentence; or evaluating self-consistency \cite{mundler2023self}, where inconsistent outputs often indicate hallucinations. Recent studies have shown that hallucinations can be attributed to the model's internal representations and have proposed white-box methods to detect hallucinations based on token latent states \cite{DBLP:conf/iclr/BurnsYKS23, DBLP:conf/acl/AzadiFD23,song2024luna,zhu-etal-2024-pollmgraph}. These approaches have outperformed black-box methods across various tasks. However, as noted in ~\citet{levinstein2024still}, they often struggle when non-factual tokens appear at the beginning or middle of the sequence.

\paragraph{Neural Differential Equations.} Neural DEs have been extensively used in modeling dynamical systems or simulating neural networks~\cite{chang2017multi, dutta2021redesigning}.  For instance, \citet{lu2018beyond} showed that any parametric ODE solver can be conceptualized as a deep learning framework with infinite depth. \citet{chen2018neural} achieved ResNet-comparable results with a drastically lower number of parameters and memory complexity by parameterizing hidden layer derivatives and using ODE solvers. In addition, \citet{lu2019understanding} was the first to draw analogies between transformers and dynamical systems, conceptualizing the transformer as a numerical approximation of ODEs. Furthermore, Neural DEs play important roles in interpolation, forecasting, and classification tasks in time series data~\cite{DBLP:conf/nips/KidgerMFL20, DBLP:conf/ijcai/LiangOYWTZ21, DBLP:conf/aistats/LiWCD20, DBLP:journals/corr/abs-2402-14989, DBLP:conf/acl/LiDZJZZXZ0Z22}.

\section{Methodology: \methodit}
\label{sec:method}

We denote the generated text as a sequence of tokens $o_{0:n}=(o_0,o_1,...,o_n)$, where $o_t$ represents the $t$-th token. Given a generated text sample $\textbf{o}=o_{0:n}$, our objective is to predict $P(\xi|\textbf{o})$ where $\xi \in \{0, 1\}$ serves as the hallucination indicator variable, with $\xi=1$ indicating a hallucination and $\xi=0$ otherwise. Naturally, each token $o_t$ is associated with an internal state representation $x_t \in \mathbb{R}^{d_x}$, derived from the specific hidden layer embeddings corresponding to token $t$. We generally use the embedding from the last layer to represent each token, where $d_x$ denotes the embedding dimension. The value of $d_x$ varies across models; for instance, $d_x=4096$ for LLama-7B, while $d_x=5120$ for LLama-13B.

\subsection{Neural DEs}
To capture the dynamic behavior of LLMs, we utilize Neural ODEs, Neural CDEs, and Neural SDEs to model the evolution in the latent space. Neural ODEs describe smooth, continuous-time dynamics using deterministic equations, Neural CDEs introduce control signals to guide system evolution. Furthermore, Neural SDEs incorporate stochasticity to account for uncertainty or noise within the system. Figure~\ref{fig:all:methods} illustrates hallucination detection using \methodit with Neural CDEs.

\paragraph{Neural ODEs.} Let $\bm{x}=x_{0:n}= (x_0,...,x_n) \in \mathbb{R}^{d_x}$ denote the embeddings in the internal space. $\bm{x}$ is projected into $\bm{y}=y_{0:n}= (y_0,...,y_n) \in \mathbb{R}^{d_y}$ by PCA. Consider a latent representation $z(t) \in \mathbb{R}^{d_z}$ at time $t$ in latent space, which is given by
\begin{equation}
\footnotesize 
\begin{split}
    &z(t) = z(0) + \int_0^t f(s, z(s);\theta_f) ds \\
    &\mbox{with} \quad z(0) = h(\bm{y};\theta_h),
    \label{eq:neural_ode}
\end{split}
\end{equation}
where $h: \mathbb{R}^{d_y} \rightarrow \mathbb{R}^{d_z}$ is an function with parameter $\theta_h$ and $f(t,z(t);\theta_f)$ is a neural network parameterized by $\theta_f$ to approximate $\frac{d \bm{z}(t)}{d t}$.  Neural ODEs rely on ODE solvers, such as the explicit Euler method~\cite{euler1845institutionum}, to solve the integral problem in~\eqref{eq:neural_ode}. Since we can freely choose the upper limit $t$ of the integration, we can predict $z$ at any time $t$. That is, once $h(\cdot;\theta_h)$ and $f(\cdot, \cdot;\theta_f)$ have been learned, then we are able to compute $\bm{z}(t)$ for any $t\geq 0$.

\paragraph{Neural CDEs.} The solution to Neural ODEs is determined by its initial condition, making it inadequate for incorporating incoming information into a differential equation. To address this issue, \citet{kidger2020neural} proposed Neural CDEs by combining a controlled path $Y(t)$ of the underlying time-series data. Specifically, given the sequential data $\bm{y}=(y_0, y_1, \ldots, y_n)$, $z(t)$ is determined by 
\begin{equation}
\footnotesize 
\begin{split}
    & z(t) = z(0) + \int_0^t f(s,z(s); \theta_f) d Y(s) \\
    & \mbox{with } \quad z(0) = h(\bm{y};\theta_h),
    \label{eq:neural_cde}
\end{split}
\end{equation}
where $Y(t)$ is chosen as a natural cubic spline path~\cite{kidger2020neural} or hermite cubic splines with backward differences~\cite{morrill2021neural} of the underlying time-series data. Differently from Neural ODEs, $f(t, z(t); \theta_f)$ is a neural network parameterized by $\theta_f$ to approximate $\frac{d z(t)}{d Y(t)}$. 


\paragraph{Neural SDEs.} 
Neural SDEs allow for describing the stochastic evolution of trace, rather than the deterministic evolution \citep{kidger2021neural,kidger2021efficient}. The latent representation $z(t)$ of Neural SDEs is governed by the following SDE:
\begin{equation}
\footnotesize
\begin{split}
    & z(t) = z(0) + \int_0^t f(s,z(s);\theta_f) d s \\   
    & + \int_0^t g(s, z(s);\theta_g) d W(s) \quad \mbox{with } \quad z(0) = h(\bm{y};\theta_h)
\label{eq:neural_sde}
\end{split}
\end{equation}
where $\{W_t\}_{t\geq 0}$ is a $d_z$-dimensional Brownian motion, $f(\cdot,\cdot;\theta_f)$ is the drift function, and $g(\cdot,\cdot;\theta_g)$ is the diffusion function. Drift and diffusion functions are represented by neural networks.


\subsection{Classifier}

We derive $z^{*}$ from the latent states $\bm{z}=(z_0, z_1, z_2, ..., z_n)$ using the function $k(\theta_k)$. The classifier $c(\theta_c)$ then classifies $z^{*}$. In this work, the classifier is implemented as a simple linear layer followed by a sigmoid function.

\subsection{DEs Solvers and Adjoint Methods}

For simplicity, we denote the parameters of neural networks used in $k(\theta_k)$, $c(\theta_c)$ and Equations~\eqref{eq:neural_ode}, \eqref{eq:neural_cde}, \eqref{eq:neural_sde} as $\bm{\theta}$. After choosing one from  Neural ODEs, Neural CDEs, or Neural SDEs to capture the state generation process in latent space, two natural questions arise:  (1) How can we generate the subsequent latent states $(z(t_1), z(t_2),...)$ based on $z(0)$ and $\bm{\theta}$?  (2) How can we update the parameters $\bm{\theta}$ to ultimately obtain the optimal solution $\bm{\theta^{*}}$? Sections \ref{sec:forward} and \ref{sec:back} answer the above two questions respectively.

\subsubsection{DE Solvers for Forward Propagation}
\label{sec:forward}
We begin by introducing two common ODE solvers: first-order and high-order schemes. Numerical methods for Neural CDEs and Neural SDEs can be adapted from these approaches.

\paragraph{First-order ODE Solvers.}
Euler method~\cite{euler1845institutionum} is the simplest method for solving ODEs. The transformation at each time step can be expressed as:
\begin{equation}
\footnotesize
z_{t+1} = z_t + \frac{dz(t)}{dt} =z_t + f(t,\bm{z}(t);\theta_f) .
\end{equation}

\paragraph{High-order ODE Solvers.}
The Euler method is not "precise" because it is a first-order method, and naturally with local truncation errors. The global error will be accumulated if we want to capture a long timestep trajectory. Herein, we use the Runge-Kutta method~\cite{runge1895numerische} for a higher-order solution to ODEs. They are a classic family of iterative methods with different orders of precision. More formally, the explicit Runge-Kutta methods of an $n$-step solution are defined to be:
\begin{equation}
\footnotesize
\begin{split}
z_{t+1} &= z_t + \sum_{i=1}^{n} \gamma_{i}Z_i, Z_1 = \Delta t f(t, z_t; \theta_f), \\
Z_i &= \Delta t f(t + \alpha_i \Delta t, z_t+\sum_{j=1}^{i-1}\beta_{ij}Z_j; \theta_f) 
\end{split}
\end{equation}

where $\Delta t$ is the time size and could be simply 1 in most cases. $Z_i$ is an intermediate approximation to the solution at step $t + \alpha_i \Delta t$. $\alpha$, $\beta$ and $\gamma$ are coefficients which can be determined by the series of $z_{t+1}$. In this work, we use fourth-order Runge-Kutta (RK4) for solving Equation~\eqref{eq:neural_ode}, details in Appendix~\ref{sec:appendix:rk4}.


 The Neural CDEs problem in \eqref{eq:neural_cde} can be solved by using the above-mentioned ODE solvers since $\frac{d z(t)}{d t} = f(t, z(t); \theta_f) \frac{d X(t)}{d t}$. However, the Neural SDE problem \eqref{eq:neural_sde} requires additional handling of stochastic noise, making its solution methods more complex. Herein, we use  the Euler-Maruyama method designed to handle noise terms, which is given by
\begin{equation}
\footnotesize
\begin{split}
z_{t+1} &=z_t + f(t,z(t);\theta_f) + g(t,z(t);\theta_f) \mathcal{Z},
\end{split}
\end{equation}
where $\mathcal{Z} \sim \mathcal{N}(0,1)$ is a standard normal random variable with mean 0 and variance 1.

\begin{table*}[hbt!]
\scriptsize
\centering
\begin{tabular}{m{1.4cm}<{\raggedright}m{0.7cm}<{\centering} m{0.7cm}<{\centering} m{0.7cm}<{\centering}m{0.7cm}<{\centering} m{0.7cm}<{\centering} m{0.8cm}<{\centering} | m{0.7cm}<{\centering} m{0.7cm}<{\centering} m{0.7cm}<{\centering} m{0.7cm}<{\centering}m{0.7cm}<{\centering} m{0.8cm}<{\centering}}
\toprule
& \multicolumn{6}{c}{$\textbf{Company}^{*}$}  & \multicolumn{6}{c}{$\textbf{Fact}^{*}$} \\
\cmidrule{2-13} 
\textbf{Method}  & LLama-2-7B     & LLama-2-13B  &Alpaca-13B   &Vicuna-13B  & Mistral-7B-0.3  & Gemma-2-9B & LLama-2-7B     & LLama-2-13B      &Alpaca-13B & Vicuna-13B &  Mistral-7B-0.3 & Gemma-2-9B \\
\midrule
\multicolumn{13}{l}{\cellcolor{lightgray}{Prompt-based Methods}}  \\
\text{P(True)} & 51.4 & 49.1  & 50.6    & 52.4 & 51.5  &  51.0  & 54.1 & 53.6 & 53.8 & 51.3 & 53.1 & 52.6    \\
\multicolumn{13}{l}{\cellcolor{lightgray}{Logit-based Methods}}  \\
\text{AvgProb} & 59.0 & 59.2  & 53.0  & 60.2  & 59.3 & 58.0 &  59.5 & 59.3  & 54.2  &  58.3 & 56.3 & 61.2    \\
\text{AvgEnt}  & 54.0 & 56.4  & 54.2  & 53.3  & 56.2 & 54.1&  54.2  & 54.1  & 50.3  & 51.2 & 53.2 & 53.4   \\
\text{EUBHD}  & 52.5 &  53.2 & 54.1  & 55.3  & 53.8  &55.6  & 59.7 & 60.8 & 59.4 & 57.9 & 56.5 & 58.2  \\
\multicolumn{13}{l}{\cellcolor{lightgray}{Classification-based Methods}} \\
SAPLMA  & 54.0  & 58.2  & 59.3  & 68.2 & 63.2& 64.8 &    58.3 & 62.4 & 59.8  &  65.5 & 59.6&  61.2  \\
\text{MIND}  & 56.4 & 60.3  &  62.4 & 69.8  &60.1  & 65.9  & 59.6  & 63.7 & 61.8& 70.7& 60.1 & 62.8    \\
\text{Probe@Exact} & 55.9 & 60.7  & 61.2 & 67.2  & 64.4 & 63.9 & 60.7 & 63.9  & 60.2 & 68.4  & 59.2 & 63.7 \\
ODEs  & 59.7   &65.3   & 67.8  & \underline{72.9} & 63.5 & 71.4 & 58.6 & 66.9 & 64.3  & 70.4  & 62.4 & 66.7 \\ 
CDEs  & \underline{65.9}   & \underline{72.8}  & \textbf{75.3}  & \textbf{79.8} & \underline{66.9} & \textbf{73.6} & \underline{67.5} & \textbf{74.8} & \textbf{72.9}  & \underline{76.7} & \underline{74.1} &  \textbf{73.9}  \\
SDEs  & \textbf{73.8}   & \textbf{78.4}  & \underline{70.5}  &  72.3 & \textbf{71.3} & \underline{72.8} & \textbf{70.3} & \underline{73.1} & \underline{70.3}  & \textbf{78.6} & \textbf{75.3} & \underline{72.5}   \\
\midrule
& \multicolumn{6}{c}{$\textbf{City}^{*}$}  & \multicolumn{6}{c}{$\textbf{Invention}^{*}$}\\
\cmidrule{2-13}
\textbf{Method}  & LLama-2-7B     & LLama-2-13B  &Alpaca-13B   &Vicuna-13B  & Mistral-7B-0.3  & Gemma-2-9B & LLama-2-7B     & LLama-2-13B      &Alpaca-13B &Vicuna-13B & Mistral-7B-0.3   &  Gemma-2-9B \\
\midrule
\text{P(True)} & 53.1 &  54.7 & 57.3  & 56.2  & 49.8 &  51.7 & 49.3  & 50.2  & 47.6  &54.7 & 51.9 & 49.7\\
\multicolumn{13}{l}{\cellcolor{lightgray}{Logit-based Methods}}  \\
\text{AvgProb}   & 54.2  & 56.3  & 51.5  & 59.2 & 53.9& 55.6  & 51.2 & 51.3 & 49.4  & 55.3 & 53.7&  52.9\\
\text{AvgEnt}    & 49.1  &50.2  & 49.3  & 52.4 & 47.1 & 52.0 & 45.1 & 46.3 & 48.4  &  47.5 & 47.1& 45.9 \\
\text{EUBHD}  &  59.9& 61.1  &  60.3 & 58.5  & 59.5 & 60.7 & 60.1 &  59.8 &  57.9 & 59.4  & 60.6 & 58.9     \\
\multicolumn{13}{l}{\cellcolor{lightgray}{Classification-based Methods}} \\
SAPLMA & 60.0  & 69.3  & 59.4  & 64.5 & 63.3 & 64.7& 59.2  & 66.0  & 52.4  & 69.3  & 61.3& 59.4  \\ 
\text{MIND}  & 64.5 & 71.3  & 62.6  &  65.8    & 63.0  &65.2   & 60.5& 65.1&53.6 & 71.2& 64.1&  58.6 \\
\text{Probe@Exact} & 65.8 & 70.4  & 61.8 & 66.9  & 62.7 & 64.3 & 61.1 & 63.0  & 55.5 & 70.2  & 63.6 & 57.3 \\
ODEs & 73.0  & \underline{82.3}  & 71.2  & 73.2 & 75.1& 72.4& 60.3 & \underline{80.9}  & 69.7 & 80.4 & \underline{79.1} & \underline{80.5} \\ 
CDEs & \underline{75.7}  & 80.6  & \underline{72.1}  & \underline{80.1} & \textbf{77.5} &\underline{77.2} & \textbf{75.9} & \textbf{88.3}  & \underline{73.8} &  \underline{81.2} & \textbf{81.3} & \textbf{83.7} \\
SDEs & \textbf{79.1}  & \textbf{89.8}  & \textbf{74.3}  & \textbf{82.5} & \underline{76.4} & \textbf{79.8} & \underline{68.7} & 79.6  & \textbf{74.2}  & \textbf{85.9} &74.3 & 79.5   \\
\bottomrule
\end{tabular}
\caption{The detection AUC-ROC (\%) of different approaches across multiple LLMs on $\text{Company}^{*}$, $\text{Fact}^{*}$, $\text{City}^{*}$ and $\text{Invention}^{*}$. \textbf{Bold} and \underline{underlined} numbers denote the best and second-best values, respectively. ODEs, CDEs, and SDEs are the abbreviations of Neural ODEs, Neural CDEs, and Neural SDEs, respectively.}
\label{tb:results:true_false}
\end{table*}

\subsubsection{Adjoint Methods for Back Propagation}
\label{sec:back}
Since Neural DEs are continuous-time models computed through DE solvers, standard backpropagation cannot be directly applied. \citet{chen2018neural} applied the adjoint sensitivity method~\cite{pontryagin2018mathematical} to compute gradients for Neural ODEs.
Specifically, to optimize the loss function $L$, we require gradients with respect to $\theta$. The first step is to determine how the gradient of the loss depends on the hidden state $z(t)$ at each instant. This quantity is called the adjoint
\begin{equation}
\footnotesize
a(t) = \frac{\partial L}{\partial z(t)}.
\end{equation}
Its dynamics are given by another ODE, which can be thought of as the instantaneous analog of the chain rule:
\begin{equation}
\footnotesize
\frac{da(t)}{dt} = -\alpha(t)^{T}\frac{\partial f(t, z(t); \theta_f)}{\partial z}.
\end{equation}
We can compute $a(t)$ by another call to an ODE solver. Computing the gradients with respect to the parameters $\theta$ requires evaluating a third integral, which depends on both $z(t)$ and $a(t)$:
\begin{equation}
\footnotesize
\frac{dL}{d\theta} = -\int_{t_1}^{t_0}\alpha(t)^{T}\frac{\partial f(t, z(t), \theta_f)}{\partial z}.
\end{equation}

In addition, \citet{DBLP:conf/nips/KidgerMFL20} and \citet{DBLP:conf/aistats/LiWCD20} proposed the adjoint sensitivity methods for Neural CDEs and Neural SDEs, respectively. In our work, we build upon the above methods to update the parameters of neural networks.

\section{Experimental Settings}

\subsection{Datasets}

\begin{table*}[hbt!]
\scriptsize
\centering
\begin{tabular}{m{1.4cm}<{\raggedright}m{0.7cm}<{\centering} m{0.7cm}<{\centering} m{0.7cm}<{\centering}m{0.7cm}<{\centering} m{0.7cm}<{\centering} m{0.8cm}<{\centering}| m{0.7cm}<{\centering} m{0.7cm}<{\centering} m{0.7cm}<{\centering} m{0.7cm}<{\centering}m{0.7cm}<{\centering} m{0.8cm}<{\centering}}
\toprule
& \multicolumn{6}{c}{\textbf{TruthfulQA}} & \multicolumn{6}{c}{\textbf{TriviaQA}} \\
\cmidrule{2-13} 
\textbf{Method}  & LLama-2-7B     & LLama-2-13B  &Alpaca-13B   &Vicuna-13B  & Mistral-7B-0.3  & Gemma-2-9B & LLama-2-7B     & LLama-2-13B      &Alpaca-13B &Vicuna-13B & Mistral-7B-0.3  &  Gemma-2-9B \\
\midrule
\multicolumn{13}{l}{\cellcolor{lightgray}{Prompt-based Methods}}  \\
\text{P(True)} & 52.5 &  53.6   & 51.3 & 54.0  & 49.7  & 50.0 & 42.3 & 44.6 & 42.1 & 50.6 & 48.3 & 49.2   \\
\multicolumn{13}{l}{\cellcolor{lightgray}{Logit-based Methods}}  \\
\text{AvgProb} & 51.4 & 54.6  & 53.3  &  55.1 &  48.3&45.6 & 44.1 &  48.3 &  43.1 & 47.1  & 48.5 & 48.0 \\
\text{AvgEnt}  & 49.4 & 53.0  & 52.7  & 53.6  & 51.0 &52.1 & 41.1 & 43.2  & 41.6  &44.5   & 47.6 & 43.5  \\
\text{EUBHD} & 81.2 & 78.1   & 77.4  & 79.7 & 80.3 & 81.4  & 80.5  & 81.1  & 78.2 & 79.1 & 80.6& 81.7   \\
\multicolumn{13}{l}{\cellcolor{lightgray}{Consistency-based Methods}}  \\
Unigram  & 57.6  & 62.2  & 60.1 & 63.4 & 60.9& 61.8& 56.8& 60.4  & 57.9  & 61.3  &59.5  &  60.3 \\
NLI         & 60.6  & 63.7  & 61.6 & 65.1  & 61.3 & 62.5& 59.4&  63.2 &58.1   & 64.5  & 61.4 & 62.1   \\
INSIDE   & 79.8 & 81.2  & 80.0  &82.1   & 81.8 & 82.4  & \underline{81.7}  &  82.6  & 78.1 & 80.8 & 81.3  & 82.0\\
\multicolumn{13}{l}{\cellcolor{lightgray}{Classification-based Methods}} \\
SAPLMA  & {87.5}  & {86.3}  & 84.9  &  88.6  & 81.3 &85.4 & 80.0 & 81.1   & 80.2  & 85.0  &84.1  & 83.4 \\
\text{MIND}  & \underline{88.0} & 87.1  &  84.5 &88.9  & 83.6 &85.7 &79.4 & 82.3  &  81.1 & 83.2  & \underline{84.5} & 81.1 \\
\text{Probe@Exact} & 85.7 & 86.8  & \underline{85.2} & 88.7  & 82.9 &\underline{87.4} &80.3 & 82.5  & \underline{81.9} & 84.4  & 84.1 & 84.0 \\
ODEs  & 84.2  & \underline{87.9}  &  83.1  &  {83.8} & 82.4 & 85.3&\underline{81.7} & \underline{83.6}  &80.5   & \underline{85.9}  & 83.7 & \underline{84.6} \\ 
CDEs  & 86.7  & {84.0}  &  {84.3} &  \underline{89.2} & \underline{83.9} & \textbf{87.7} & \textbf{83.7} & \textbf{84.9}  & \textbf{82.6}  & \textbf{86.3}  & 84.1 & \textbf{85.0} \\
SDEs  & \textbf{88.3}  & \textbf{89.3}  &  \textbf{86.4} & \textbf{89.5} & \textbf{85.1} & 87.0 & 81.0 & 83.3  &  81.5 & 84.3  & \textbf{85.1} & 83.2 \\
\midrule
& \multicolumn{6}{c}{\textbf{HaluEval}} & \multicolumn{6}{c}{\textbf{NQ}} \\
\cmidrule{2-13} 
\textbf{Method}  & LLama-2-7B     & LLama-2-13B  &Alpaca-13B   &Vicuna-13B  & Mistral-7B-0.3  & Gemma-2-9B  & LLama-2-7B     & LLama-2-13B      &Alpaca-13B &Vicuna-13B & Mistral-7B-0.3  &  Gemma-2-9B \\
\midrule
\multicolumn{13}{l}{\cellcolor{lightgray}{Prompt-based Methods}}  \\
\text{P(True)} &46.7  & 48.9  & 51.6  & 50.2  & 49.7 &  46.8  &54.7  & 56.8 & 51.0 & 53.4 & 52.1 & 51.3   \\
\multicolumn{13}{l}{\cellcolor{lightgray}{Logit-based Methods}}  \\
\text{AvgProb} &42.1  & 44.4  & 43.2  &45.7   & 43.6 &44.5 & 54.3& 55.9  &53.1   &56.4   &54.7  & 55.3 \\
\text{AvgEnt}  &47.3  & 48.5  & 46.1  & 51.4  & 49.7 &50.3 &53.9 & 54.6  & 54.2  & 55.2  & 53.8 & 54.6 \\
\text{EUBHD} &71.9  & 78.1  &71.3  & 76.0  &70.5  &72.6   & 73.9  & 79.4 & 76.8 & 73.2 & 71.7 & 70.3    \\
\multicolumn{13}{l}{\cellcolor{lightgray}{Consistency-based Methods}}  \\
Unigram    & 58.2 & 57.1  & 57.9  & 57.6 & 62.3& 59.4 &63.1 & 65.2  & 62.9  & 67.8  &64.5  & 65.1   \\
NLI         &61.3  & 60.2  & 55.2  &  62.5 & 63.1 & 64.4& 64.2 &  66.9 & 63.8  & 65.4  &62.6  &  64.0  \\
INSIDE   & 74.5 & 76.9  &  73.3 & 75.2  & 76.0 & 75.8  &76.8   & 77.1 & 74.3 & 75.8  &76.4  & 75.9\\
\multicolumn{13}{l}{\cellcolor{lightgray}{Classification-based Methods}} \\
SAPLMA  & 87.0 & 90.1  & 89.5  &93.1   & 89.4& 90.5&89.1 & 90.5 & 87.6  & \underline{93.2}  &\underline{90.3}  & 88.9\\
\text{MIND}  &86.1  &93.8   & 93.7  & 92.9 &\underline{94.5} &91.0 &90.5 & \underline{93.6} & 92.7  & 90.6  & 87.2 & 89.5  \\
\text{Probe@Exact} & 88.3 & 92.4  & 93.5 & 94.1  & 93.4 & 92.1 & 92.0 & 91.9  & \underline{92.8} & 92.3  & 88.6 & 90.3 \\
ODEs  & 89.5 & 93.9  & 92.1 & \underline{95.4} &91.2  &90.5 & 91.3& 92.1  & 90.5  & 92.4  & 89.7 & 90.0\\ 
CDEs  & \underline{91.4}  & \textbf{97.1} & \underline{95.3} &\textbf{96.9}  & \textbf{95.4} & \textbf{96.0} &\underline{93.7} & \textbf{95.2}  & 92.1  & \textbf{93.6}  & \textbf{90.5} & \textbf{91.8} \\
SDEs  & \textbf{92.8} & \underline{95.4} &\textbf{97.1}  & 93.1& 93.7&\underline{92.6} & \textbf{94.1} & 93.2  & \textbf{93.5}  & 91.1  &89.7  & \underline{90.9} \\
\bottomrule
\end{tabular}
\caption{The detection AUC-ROC (\%) for different approaches over multiple LLMs on TruthfulQA, TriviaQA, HaluEval and NQ.}
\label{tb:results:truthful_qa}
\end{table*}

\paragraph{True-False Dataset.} The original dataset consists of six sub-datasets, each named after its subject matter~\cite{azaria-mitchell-2023-internal}. We follow the method proposed in~\citet{levinstein2024still} to create factual and non-factual statements containing subtle differences. Specifically, we prompt GPT-4o to generate new statements that are factually opposite to the original while maintaining only minor word differences. For example, we obtain a non-factual statement "The earth doesn't orbit the sun." from the factual statement "The earth orbits the sun." For our experiments, we randomly select 550, 560, 500, and 500 statements from the \textit{Companies}, \textit{Scientific Facts}, \textit{Cities}, and \textit{Inventions} sub-datasets, respectively. The resulting datasets are referred to as $\textit{Company}^{*}$, $\textit{Fact}^{*}$, $\textit{City}^{*}$, and $\textit{Invention}^{*}$. This dataset poses a greater challenge for hallucination detection.

\paragraph{Question Answering Datasets.} We utilize four widely used question answering datasets, including \textit{TruthfulQA}~\cite{lin-etal-2022-truthfulqa}, \textit{TriviaQA}~\cite{joshi-etal-2017-triviaqa}, "QA" subset of \textit{HaluEval}~\cite{li-etal-2023-halueval}  and \textit{NQ}~\cite{kwiatkowski-etal-2019-natural}. Each question is accompanied by a truthful and a hallucinatory answer. Unlike the True-False dataset, we use the Levenshtein~\cite{levenshtein1966binary} distance to select the pair of correct and incorrect answers with the greatest textual difference. These pairs, along with the original questions, form the data used for our experiments. Finally, we generate 1,000 samples in each of the four aforementioned datasets. 



\subsection{ Models} 
We evaluate both our method and baseline approaches using common open-source LLMs, including LLama-2-7B, LLama-2-13B~\cite{touvron2023llama}, Alpaca-13B~\cite{alpaca}, Vicuna-13B-v1.3~\cite{vicuna2023}, Mistral-7B-v0.3~\cite{jiang2023mistral} and Gemma-2-9B~\cite{team2024gemma}. 



\subsection{Baselines} 
\label{sec:baseline}
We choose the following four types of hallucination detection methods as baselines. More details are shown in Appendix~\ref{sec:app_baseline}.

\paragraph{Prompt-based methods} utilize a simple prompt template to enable the model to assess the correctness of the response. Here, we use \textit{P(True)}, proposed in \citet{kadavath2022language}, as a representative of this class of methods.

\paragraph{Logit-based methods} use the uncertainty of LLMs' outputs to detect hallucination. We adopt the two effective metrics used in~\citet{huang2023look}, namely \emph{\text{AvgProb}}, \emph{\text{AvgEnt}}, to aggregate logit-based uncertainty of all tokens to measure sentence uncertainty. In addition, we also compare our approach with \textit{EUBHD}~\cite{zhang-etal-2023-enhancing-uncertainty}, which focuses on key tokens rather than considering all tokens. 


\begin{figure}[hbt!]
    \centering
    \begin{subfigure}[b]{0.48\linewidth}
    \includegraphics[width=\textwidth]{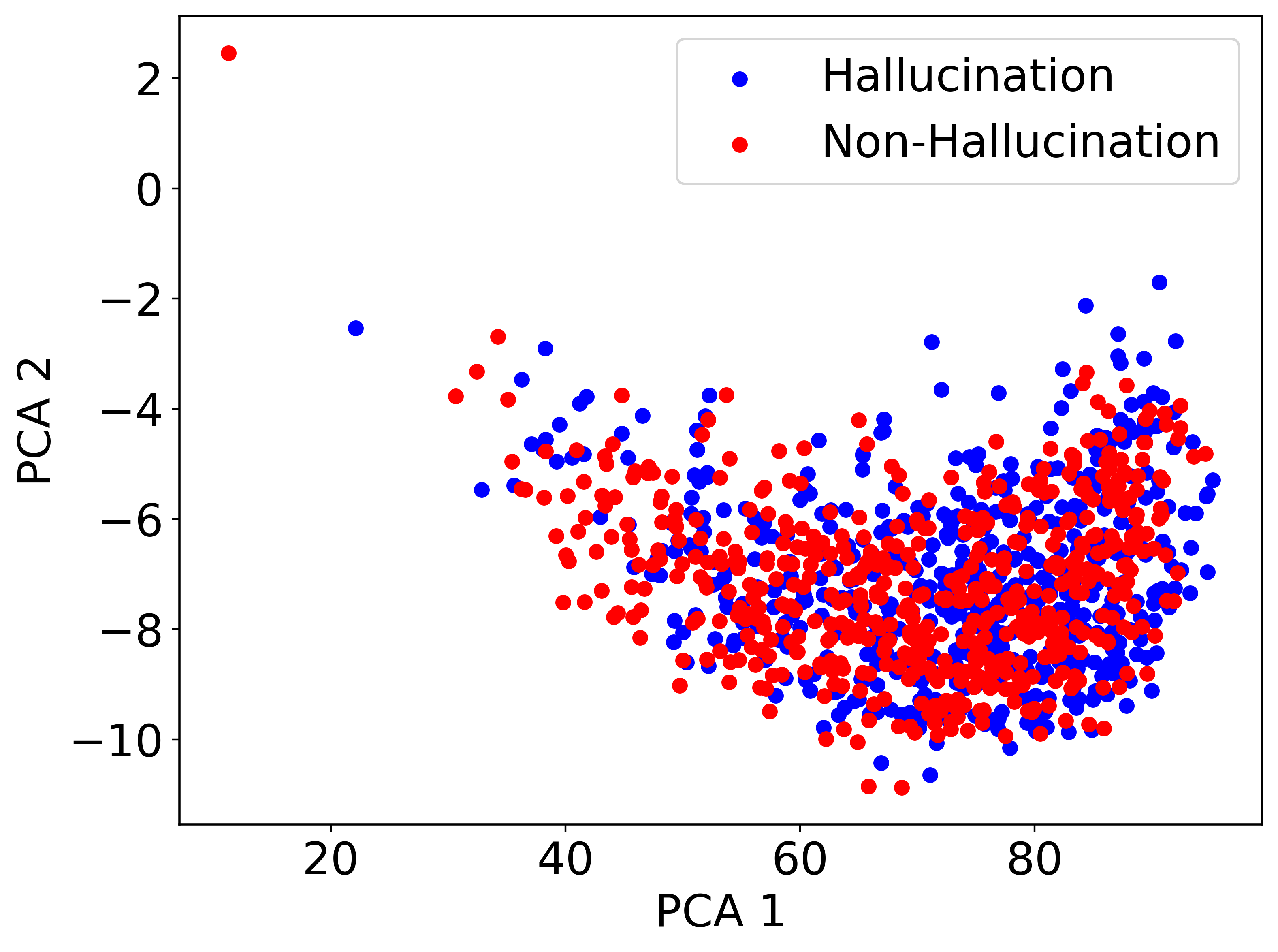}
    \caption{$\text{Fact}^{*}$}
    \label{fig:result:logp}
    \end{subfigure}
    \hspace{.00002in}
    \begin{subfigure}[b]{0.48\linewidth}
        \includegraphics[width=\textwidth]{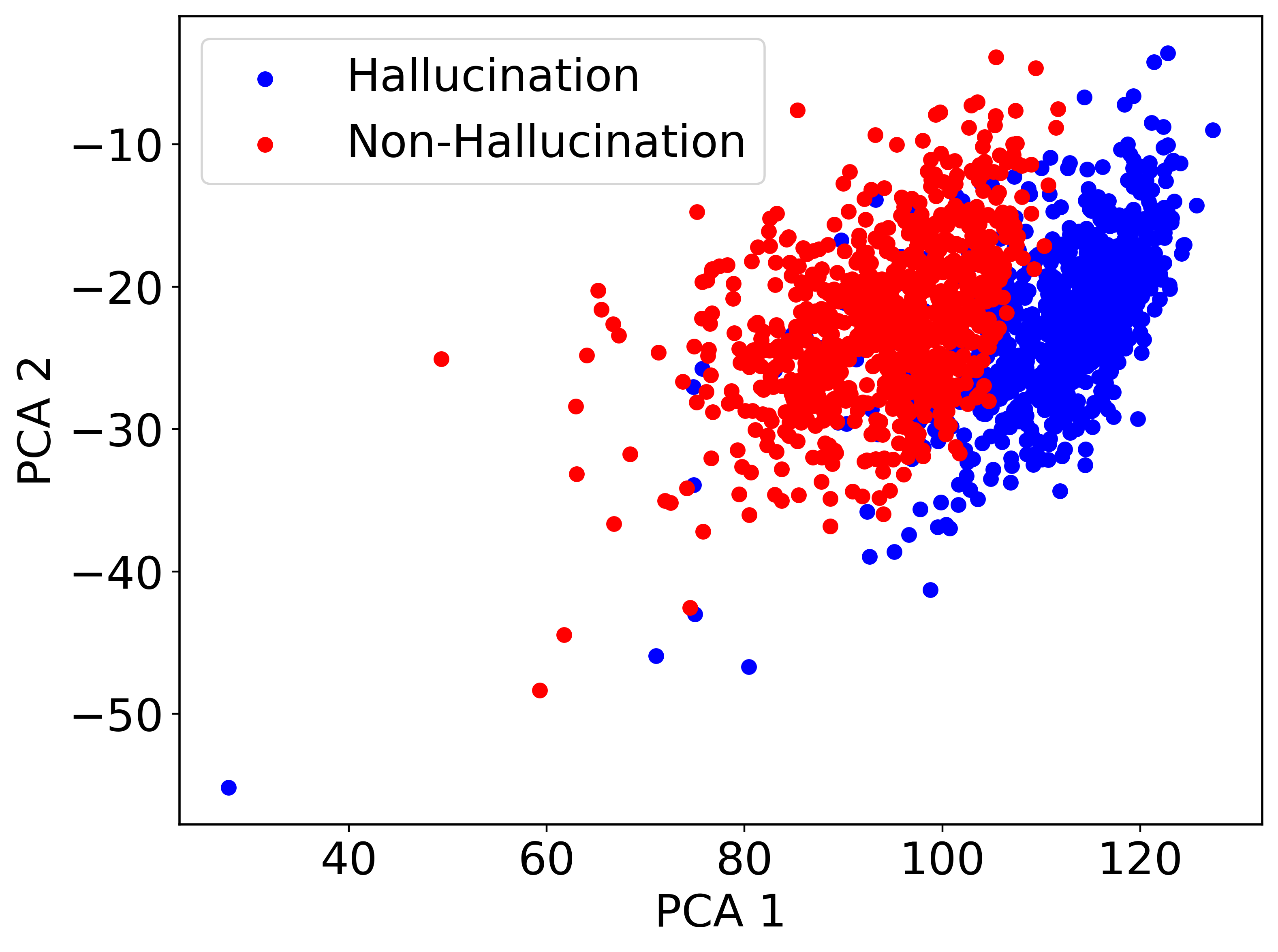}
    \caption{$\text{TruthfulQA}$}
    \label{fig:result:probing}
    \end{subfigure}
    \caption{2D PCA projection of the last hidden layer's embedding for the final token on $\text{Fact}^{*}$ and $\text{TruthfulQA}$. Blue and red dots represent hallucinations and non-hallucinations, respectively.}
    \label{fig:hidden_information:tr_false}
\end{figure}

\paragraph{Consistency-based methods} are motivated by the idea that if an LLM possesses specific knowledge, the sampled responses are likely to be similar and contain consistent facts. In this work, we apply two important variants proposed in~\citet{manakul2023selfcheckgpt}, namely \emph{Unigram} and \emph{Natural Language Inference} (NLI), as well as \textit{INSIDE} by~\citet{chen2024inside}, which leverages the eigenvalues of the covariance matrix of responses.

\paragraph{Classification-based methods} train a classifier on a dataset containing labeled statements. We choose \textit{SAPLMA}~\cite{azaria-mitchell-2023-internal}, \textit{MIND}~\cite{su-etal-2024-unsupervised} and \textit{Probe@Exact}~\cite{orgad2025llms} as representatives of this type of method. Unlike SAPLMA, which relies on pre-annotated datasets, MIND automatically labels data during the detection process to train its classifier. SAPLMA utilizes information from the last token, whereas Probe@Exact relies on information from potential correct tokens.


\subsection{Evaluation Metric}
We utilize \textit{AUC-ROC}, which stands for the area under the ROC curve, to objectively evaluate the effectiveness of models. The higher value of AUC-ROC, the stronger the ability of this method for hallucination detection. All experiments are conducted on NVIDIA A100 GPUs with 40GB of memory.

\subsection{Implementation Details}
\label{sec:implementation_details}

\paragraph{\methodit.} To reduce computational complexity, we employ PCA to reduce the dimensionality of the internal space to $K=1024$. The integrands $h(\cdot; \theta_h)$, $f(\cdot, \cdot; \theta_f)$, $g(\cdot, \cdot; \theta_g)$  in Equations~\eqref{eq:neural_ode},~\eqref{eq:neural_cde} and~\eqref{eq:neural_sde} are taken to be feedforward neural networks. Specifically, we use a single hidden layer network to represent $h(\cdot; \theta_h)$ in all variants of our methods. We use an 8-layer neural network to represent $f(\cdot, \cdot; \theta_f)$ in Neural CDEs and 10-layer neural networks for $f(\cdot, \cdot; \theta_f)$ in both Neural ODEs and Neural SDEs. Additionally, $g(\cdot, \cdot; \theta_g)$ in Neural SDEs is represented by a 4-layer neural network. A final linear layer is always applied to map the latent state to the output.  We use ReLU activation functions for Neural CDEs and Neural SDEs, while tanh activations are used for Neural ODEs. The binary cross-entropy loss is applied to the sigmoid of the model output. Additionally, we employ the Adam optimizer with a learning rate of 0.001, a batch size of 32, and 50 epochs.



\paragraph{Classification-based methods.} The classifier receives embeddings from the last layer of LLMs. In ablation studies, we discuss the results of using information from the middle layers. Different classifiers are used for different methods. More implementation details are introduced in Appendix~\ref{sec:appendix_implements}.


\section{Experimental Results and Analysis}

\subsection{Effectiveness of \methodit} 
 
\paragraph{True-False Dataset.} The comprehensive results are demonstrated in  Table \ref{tb:results:true_false}. Since consistency-based methods rely on question-and-answer pairs, and the True-False dataset is not structured in this format, we do not include this type of method as a comparison in this dataset. \textbf{It is obvious that our methods surpass SAPLMA, MIND, and Probe@Exact by a noticeable margin, evidenced by an average increase of over 14\% in the detection of AUC-ROC across different models and subsets.} Particularly, Neural CDEs outperform SAPLMA by 24.3\% on $\text{Invention}^{*}$ when using Gemma-2-9B. Even in the worst case, Neural ODEs perform comparably to SAPLMA on $\text{Fact}^{*}$ based on LLama-2-7B. Furthermore, in most cases, prompt-based methods and logit-based methods perform worse than the classification-based methods.

For different variants of our approach, \textbf{we can find that \cde and \sde outperform \ode.} As shown in Table \ref{tb:results:true_false}, the best and second-best values are achieved by \cde and \sde models in 19 out of 24 cases. The likely reason is that \cde and \sde can capture richer dynamical behaviors than \ode. As mentioned in Section~\ref{sec:method}, \cde incorporates control theory, enabling the dynamic system to account for the influence of incoming information, and \sde introduces stochasticity into the modeling process.  While \ode assumes deterministic dynamics, which can limit its flexibility in modeling the dynamics of LLMs.


\begin{figure}[t] 
\centering
\includegraphics[width=0.90\linewidth]{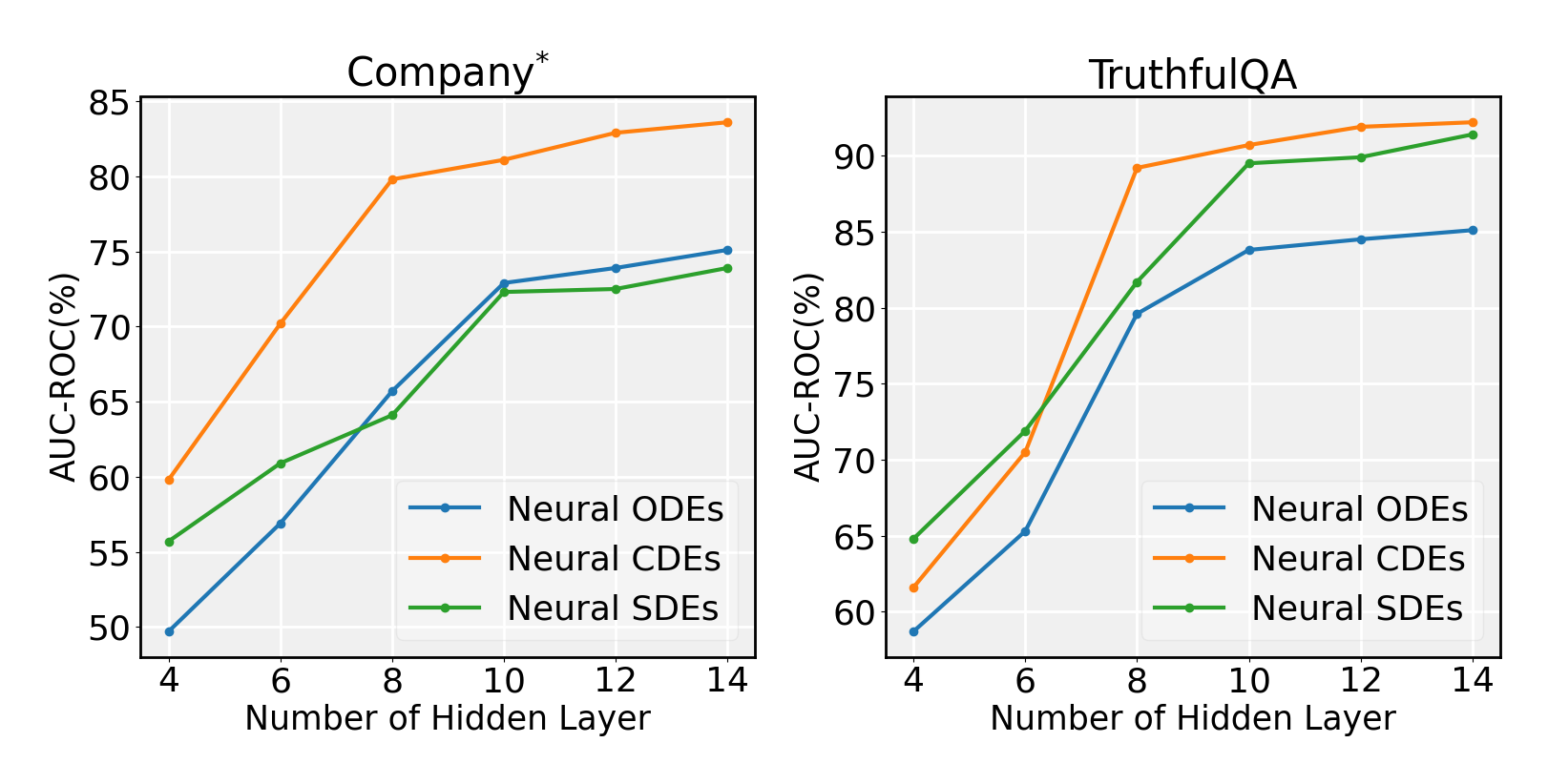} 
\caption{The impact of the number of hidden layers on Vicuna-13B: $\text{Company}^{*}$ and TruthfulQA.}
\label{fig:layer_number}
\end{figure}


\paragraph{Question Answering Datasets.} Table~\ref{tb:results:truthful_qa} shows the results on question answering datasets. \textbf{EUBHD, SAPLMA, MIND, and Probe@Exact demonstrate significantly better performance on the four question answering datasets compared to the True-False dataset across all six models.} Notably, SAPLMA, MIND and and Probe@Exact achieve comparable performance to \methodit, including \ode, \cde, and \sde, with a difference of less than 6\%. Specifically, SAPLMA outperforms \ode and \cde on TruthfulQA using LLama-2-7B and Alpaca-13B, while remaining slightly behind \sde. On the NQ dataset, MIND and Probe@Exact achieve the second-highest performance among all methods on LLama-2-13B and Alpaca-13B, respectively. Meanwhile, INSIDE ranks just below \cde on TriviaQA with LLama-2-7B.

\subsection{Analysis}

We try to understand why \methodit obviously outperforms SAPLMA, and MIND on the True-False dataset, yet performs comparably to them on the question-answer datasets. We use the subsets $\text{Fact}^{*}$ from True-False and $\text{TruthfulQA}$ as examples. We employ PCA to reduce the dimensions of the hidden embeddings, retaining the two dominant components. The results are shown in Figure~\ref{fig:hidden_information:tr_false}. The 2D PCA projection reveals a significant overlap between correct and incorrect statements in True-False, with many points intertwined. The poor separation causes other methods to perform only marginally better than random guessing in many cases. In contrast, the 2D PCA projections of TruthfulQA reveal a much clearer distinction between hallucination and non-hallucination. The statements in TruthfulQA exhibit substantial variation, as we use the Levenshtein distance to select statements with significant differences. This allows other baselines to more easily differentiate them based on the embeddings of the final token. Appendix~\ref{sec:appendix_more_dataset} contains more results on other datasets. 



\begin{figure}[t] 
\centering
\includegraphics[width=0.90\linewidth]{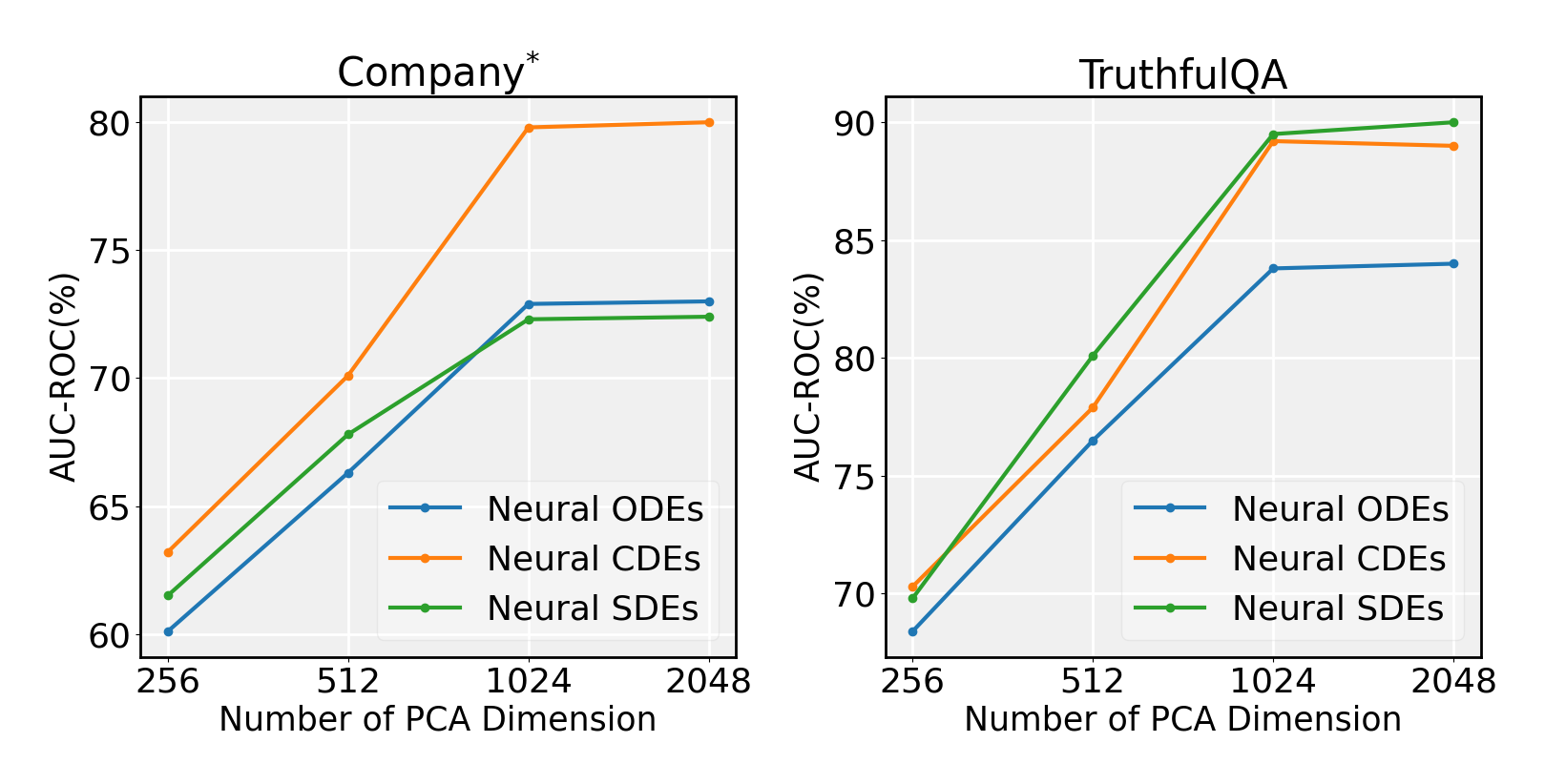} 
\caption{The impact of the PCA projection dimensions on Vicuna-13B: $\text{Company}^{*}$ and TruthfulQA.}
\label{fig:dimension_number}
\end{figure}


\subsection{Ablation Studies}



\paragraph{Number of Hidden Layers.} An important factor impacting the performance of detection methods is the number of hidden layers in neural networks representing  $f(s, z(s); \theta_f)$. Results are shown in Figure~\ref{fig:layer_number}.
Specifically, the performance of Neural CDEs improves substantially as the number of layers increases up to 8, with further increases beyond 8 still showing gains but at a slower pace. For Neural ODEs and Neural SDEs, this turning point occurs when the layer number reaches 10, based on the results from both datasets.
Finally, we ultimately set 8 layers for Neural CDEs and 10 layers for both Neural ODEs and Neural SDEs in Section~\ref{sec:implementation_details}.



\begin{table*}[hbt!]
\small
\centering
\begin{tabular}{m{1.3cm}<{\centering}!{\color{black!90}\vrule}m{1.3cm}<{\centering}m{1.3cm}<{\centering} m{1.7cm}<{\centering} m{1.3cm}<{\centering}m{1.3cm}<{\centering} m{1.3cm}<{\centering}}
\toprule
& SAPLMA & MIND & Probe@Exact & ODEs & CDEs & SDEs \\
\midrule 
 16th  & 69.4 & 70.0 & 66.7 & 72.3 & 80.0 & 73.6\\
 \cellcolor{lightgray!50} 20th  & \cellcolor{lightgray!50} 70.5 & \cellcolor{lightgray!50} 70.3 & \cellcolor{lightgray!50} \textbf{69.1} & \cellcolor{lightgray!50} 73.9 & \cellcolor{lightgray!50} \textbf{81.7} & \cellcolor{lightgray!50} 73.8\\
 24th  & \textbf{71.0} & \textbf{71.3} & 68.8 & \textbf{74.0} & 80.2 & 72.4\\
 \cellcolor{lightgray!50} 28th  & \cellcolor{lightgray!50} 68.1 & \cellcolor{lightgray!50} 68.9 & \cellcolor{lightgray!50} 67.5 & \cellcolor{lightgray!50} 72.4 & \cellcolor{lightgray!50} 78.9 & \cellcolor{lightgray!50} \textbf{74.0}\\
 Last  & 68.2 & 69.8 & 67.2 & 72.9 & 79.8 & 72.3\\
 \bottomrule
 \end{tabular}
 \caption{AUC-ROC(\%) for detection across the 16th, 20th, 24th, and 28th layers for different approaches.}
 \label{tab:middle_layers}
\end{table*}

\begin{table*}[hbt!]
\small
\centering
\begin{tabular}{m{2.5cm}<{\centering}!{\color{black!90}\vrule}m{1.3cm}<{\centering}m{1.3cm}<{\centering} m{1.7cm}<{\centering} m{1.3cm}<{\centering}m{1.3cm}<{\centering} m{1.3cm}<{\centering}}
\toprule
Methods & SAPLMA & MIND & Probe@Exact & ODEs & CDEs & SDEs \\
\midrule 
AUC-ROC (\%) & 65.4 & 67.6 & 69.4 & 79.6 & 80.1 & 84.3 \\
\bottomrule
\end{tabular}
\caption{AUC-ROC (\%) for detection on $\text{Invention}^{*}$ using classifiers trained on $\text{Company}^{*}$, $\text{Fact}^{*}$, and $\text{City}^{*}$.}
\label{tab:ood}
\end{table*}

\paragraph{Dimensions in Latent Space.} Another key factor is the dimension of the latent state after being mapped from the internal space to the latent space by PCA. We then examine the impact of varying dimensions as shown in Figure~\ref{fig:dimension_number}. \textbf{An evident improvement in detection effectiveness is associated with retaining more components during down-projection.} Therefore, all three variants of our methods achieve the best performance on both datasets when the dimension is set to 1024, affirming our hyperparameter setting in Section~\ref{sec:implementation_details}.


\paragraph{Experiment of Using Middle Layers.} We select the 16th, 20th, 24th, and 28th layers as representative intermediate layers and evaluate the performance of various methods based on the Vicuna-13B-v1.3 model on the $\text{Company}^{*}$ dataset, as shown in Table~\ref{tab:middle_layers}. Compared to the final layer, the results at the 20th and 24th layers show an overall improvement. For other layers, the results vary depending on the method. Therefore, specific intermediate layers may contain more information for whether a hallucination is occurring.

\paragraph{Experiment of the Out-of-Domain Setting.} To evaluate the generalization capability of the proposed method in an out-of-domain setting, we train the model on $\text{Company}^{*}$, $\text{Fact}^{*}$, and $\text{City}^{*}$, and test its performance on $\text{Invention}^{*}$, based on the Vicuna-13B-v1.3 model. The detailed experimental results are shown in Table~\ref{tab:ood}. All methods exhibit a certain degree of performance degradation. Compared to SAPLMA, MIND, and Probe@Exact, the proposed ODEs, CDEs, and SDEs demonstrate relatively smaller declines, with reductions of less than 2\%.







\section{Conclusion}
In this paper, we introduce \methodit, which tracks the dynamic changes in latent space. \methodit can effectively detect logical or factual inconsistencies that arise in the generated text. Comprehensive empirical results demonstrate that our approach surpasses various state-of-the-art methods by over 14\% on the True-False Dataset. 


\section*{Limitations}
This work identifies four major limitations. First, the model's training process is approximately twice as long as that of the SAPLMA method. Second, NeuralDEs, as currently presented, do not provide uncertainty estimates for their predictions, though such extensions may be feasible in the future. Third, we experiment with a limited set of numerical schemes, and other methods could potentially exploit the structure of differential equations to further improve performance. Fourth, the proposed method relies on internal activations and is therefore suited for hallucination detection in open-source models.


\section*{Ethics and Broader Impact}
We sampled a portion of the data from existing datasets for our experiments, which may affect the accuracy of some of our conclusions.

\bibliography{custom_update_update}

\begin{thebibliography}{57}
\providecommand{\natexlab}[1]{#1}

\bibitem[{Abdi and Williams(2010)}]{abdi2010principal}
Herv{\'e} Abdi and Lynne~J Williams. 2010.
\newblock Principal component analysis.
\newblock \emph{Wiley interdisciplinary reviews: computational statistics}, 2(4):433--459.

\bibitem[{Azadi et~al.(2023)Azadi, Faili, and Dousti}]{DBLP:conf/acl/AzadiFD23}
Fatemeh Azadi, Heshaam Faili, and Mohammad~Javad Dousti. 2023.
\newblock \href {https://doi.org/10.18653/v1/2023.findings-acl.782} {{PMI}-align: Word alignment with point-wise mutual information without requiring parallel training data}.
\newblock In \emph{Findings of the Association for Computational Linguistics: ACL 2023}, pages 12366--12377, Toronto, Canada. Association for Computational Linguistics.

\bibitem[{Azaria and Mitchell(2023)}]{azaria-mitchell-2023-internal}
Amos Azaria and Tom Mitchell. 2023.
\newblock \href {https://doi.org/10.18653/v1/2023.findings-emnlp.68} {The internal state of an {LLM} knows when it{'}s lying}.
\newblock In \emph{Findings of the Association for Computational Linguistics: EMNLP 2023}, pages 967--976, Singapore. Association for Computational Linguistics.

\bibitem[{Baier-Reinio and De~Sterck(2020)}]{baier2020n}
Aaron Baier-Reinio and Hans De~Sterck. 2020.
\newblock \href {https://arxiv.org/abs/2010.11358} {N-ode transformer: A depth-adaptive variant of the transformer using neural ordinary differential equations}.
\newblock \emph{arXiv preprint arXiv:2010.11358}.

\bibitem[{Burns et~al.(2023)Burns, Ye, Klein, and Steinhardt}]{DBLP:conf/iclr/BurnsYKS23}
Collin Burns, Haotian Ye, Dan Klein, and Jacob Steinhardt. 2023.
\newblock \href {https://openreview.net/pdf?id=ETKGuby0hcs} {Discovering latent knowledge in language models without supervision}.
\newblock In \emph{The Eleventh International Conference on Learning Representations, {ICLR} 2023, Kigali, Rwanda, May 1-5, 2023}. OpenReview.net.

\bibitem[{Chang et~al.(2018)Chang, Meng, Haber, Tung, and Begert}]{chang2017multi}
Bo~Chang, Lili Meng, Eldad Haber, Frederick Tung, and David Begert. 2018.
\newblock \href {https://openreview.net/forum?id=SyJS-OgR-} {Multi-level residual networks from dynamical systems view}.
\newblock In \emph{6th International Conference on Learning Representations, {ICLR} 2018, Vancouver, BC, Canada, April 30 - May 3, 2018, Conference Track Proceedings}. OpenReview.net.

\bibitem[{Chen et~al.(2024)Chen, Liu, Chen, Gu, Wu, Tao, Fu, and Ye}]{chen2024inside}
Chao Chen, Kai Liu, Ze~Chen, Yi~Gu, Yue Wu, Mingyuan Tao, Zhihang Fu, and Jieping Ye. 2024.
\newblock \href {https://openreview.net/forum?id=Zj12nzlQbz} {{INSIDE}: {LLM}s' internal states retain the power of hallucination detection}.
\newblock In \emph{The Twelfth International Conference on Learning Representations}.

\bibitem[{Chen et~al.(2018)Chen, Rubanova, Bettencourt, and Duvenaud}]{chen2018neural}
Tian~Qi Chen, Yulia Rubanova, Jesse Bettencourt, and David Duvenaud. 2018.
\newblock \href {https://proceedings.neurips.cc/paper/2018/hash/69386f6bb1dfed68692a24c8686939b9-Abstract.html} {Neural ordinary differential equations}.
\newblock In \emph{Advances in Neural Information Processing Systems 31: Annual Conference on Neural Information Processing Systems 2018, NeurIPS 2018, December 3-8, 2018, Montr{\'{e}}al, Canada}, pages 6572--6583.

\bibitem[{Chiang et~al.(2023)Chiang, Li, Lin, Sheng, Wu, Zhang, Zheng, Zhuang, Zhuang, Gonzalez, Stoica, and Xing}]{vicuna2023}
Wei-Lin Chiang, Zhuohan Li, Zi~Lin, Ying Sheng, Zhanghao Wu, Hao Zhang, Lianmin Zheng, Siyuan Zhuang, Yonghao Zhuang, Joseph~E. Gonzalez, Ion Stoica, and Eric~P. Xing. 2023.
\newblock \href {https://lmsys.org/blog/2023-03-30-vicuna/} {Vicuna: An open-source chatbot impressing gpt-4 with 90\%* chatgpt quality}.

\bibitem[{Choi et~al.(2022)Choi, Choi, Hwang, and Park}]{choi2022graph}
Jeongwhan Choi, Hwangyong Choi, Jeehyun Hwang, and Noseong Park. 2022.
\newblock \href {https://ojs.aaai.org/index.php/AAAI/article/view/20587} {Graph neural controlled differential equations for traffic forecasting}.
\newblock In \emph{Thirty-Sixth {AAAI} Conference on Artificial Intelligence, {AAAI} 2022, Thirty-Fourth Conference on Innovative Applications of Artificial Intelligence, {IAAI} 2022, The Twelveth Symposium on Educational Advances in Artificial Intelligence, {EAAI} 2022 Virtual Event, February 22 - March 1, 2022}, pages 6367--6374. {AAAI} Press.

\bibitem[{Duan et~al.(2023)Duan, Cheng, Wang, Wang, Zavalny, Xu, Kailkhura, and Xu}]{duan2023shifting}
Jinhao Duan, Hao Cheng, Shiqi Wang, Chenan Wang, Alex Zavalny, Renjing Xu, Bhavya Kailkhura, and Kaidi Xu. 2023.
\newblock \href {https://arxiv.org/abs/2307.01379} {Shifting attention to relevance: Towards the uncertainty estimation of large language models}.
\newblock \emph{ArXiv preprint}, abs/2307.01379.

\bibitem[{Dutta et~al.(2021)Dutta, Gautam, Chakrabarti, and Chakraborty}]{dutta2021redesigning}
Subhabrata Dutta, Tanya Gautam, Soumen Chakrabarti, and Tanmoy Chakraborty. 2021.
\newblock \href {https://proceedings.neurips.cc/paper/2021/hash/2bd388f731f26312bfc0fe30da009595-Abstract.html} {Redesigning the transformer architecture with insights from multi-particle dynamical systems}.
\newblock In \emph{Advances in Neural Information Processing Systems 34: Annual Conference on Neural Information Processing Systems 2021, NeurIPS 2021, December 6-14, 2021, virtual}, pages 5531--5544.

\bibitem[{Euler(1845)}]{euler1845institutionum}
Leonhard Euler. 1845.
\newblock \emph{Institutionum calculi integralis}, volume~4.
\newblock impensis Academiae imperialis scientiarum.

\bibitem[{Geng et~al.(2023)Geng, Cai, Wang, Koeppl, Nakov, and Gurevych}]{geng2023survey}
Jiahui Geng, Fengyu Cai, Yuxia Wang, Heinz Koeppl, Preslav Nakov, and Iryna Gurevych. 2023.
\newblock \href {https://arxiv.org/abs/2311.08298} {A survey of language model confidence estimation and calibration}.
\newblock \emph{ArXiv preprint}, abs/2311.08298.

\bibitem[{He et~al.(2023)He, Gao, and Chen}]{DBLP:conf/iclr/HeGC23}
Pengcheng He, Jianfeng Gao, and Weizhu Chen. 2023.
\newblock \href {https://openreview.net/pdf?id=sE7-XhLxHA} {Debertav3: Improving deberta using electra-style pre-training with gradient-disentangled embedding sharing}.
\newblock In \emph{The Eleventh International Conference on Learning Representations, {ICLR} 2023, Kigali, Rwanda, May 1-5, 2023}. OpenReview.net.

\bibitem[{Huang et~al.(2023)Huang, Song, Wang, Chen, and Ma}]{huang2023look}
Yuheng Huang, Jiayang Song, Zhijie Wang, Huaming Chen, and Lei Ma. 2023.
\newblock \href {https://arxiv.org/abs/2307.10236} {Look before you leap: An exploratory study of uncertainty measurement for large language models}.
\newblock \emph{ArXiv preprint}, abs/2307.10236.

\bibitem[{James et~al.(2021)James, Patrick, Lingyi, and Terry}]{morrill2021neural}
Morrill James, Kidger Patrick, Yang Lingyi, and Lyons Terry. 2021.
\newblock \href {https://arxiv.org/abs/2106.11028} {Neural controlled differential equations for online prediction tasks}.
\newblock \emph{ArXiv preprint}, abs/2106.11028.

\bibitem[{Jhin et~al.(2024)Jhin, Shin, Kim, Hong, Jo, Park, Park, Lee, Maeng, and Jeon}]{jhin2024attentive}
Sheo~Yon Jhin, Heejoo Shin, Sujie Kim, Seoyoung Hong, Minju Jo, Solhee Park, Noseong Park, Seungbeom Lee, Hwiyoung Maeng, and Seungmin Jeon. 2024.
\newblock \href {https://arxiv.org/pdf/2109.01876} {Attentive neural controlled differential equations for time-series classification and forecasting}.
\newblock \emph{Knowledge and Information Systems}, 66(3):1885--1915.

\bibitem[{Jiang et~al.(2023)Jiang, Sablayrolles, Mensch, Bamford, Chaplot, Casas, Bressand, Lengyel, Lample, Saulnier et~al.}]{jiang2023mistral}
Albert~Q Jiang, Alexandre Sablayrolles, Arthur Mensch, Chris Bamford, Devendra~Singh Chaplot, Diego de~las Casas, Florian Bressand, Gianna Lengyel, Guillaume Lample, Lucile Saulnier, et~al. 2023.
\newblock \href {https://arxiv.org/pdf/2310.06825} {Mistral 7b}.
\newblock \emph{arXiv preprint arXiv:2310.06825}.

\bibitem[{Joshi et~al.(2017)Joshi, Choi, Weld, and Zettlemoyer}]{joshi-etal-2017-triviaqa}
Mandar Joshi, Eunsol Choi, Daniel Weld, and Luke Zettlemoyer. 2017.
\newblock \href {https://doi.org/10.18653/v1/P17-1147} {{T}rivia{QA}: A large scale distantly supervised challenge dataset for reading comprehension}.
\newblock In \emph{Proceedings of the 55th Annual Meeting of the Association for Computational Linguistics (Volume 1: Long Papers)}, pages 1601--1611, Vancouver, Canada. Association for Computational Linguistics.

\bibitem[{Kadavath et~al.(2022)Kadavath, Conerly, Askell, Henighan, Drain, Perez, Schiefer, Hatfield-Dodds, DasSarma, Tran-Johnson et~al.}]{kadavath2022language}
Saurav Kadavath, Tom Conerly, Amanda Askell, Tom Henighan, Dawn Drain, Ethan Perez, Nicholas Schiefer, Zac Hatfield-Dodds, Nova DasSarma, Eli Tran-Johnson, et~al. 2022.
\newblock \href {https://arxiv.org/abs/2207.05221} {Language models (mostly) know what they know}.
\newblock \emph{arXiv preprint arXiv:2207.05221}.

\bibitem[{Kidger(2022)}]{kidger2022neural}
Patrick Kidger. 2022.
\newblock \href {https://arxiv.org/abs/2202.02435} {On neural differential equations}.
\newblock \emph{ArXiv preprint}, abs/2202.02435.

\bibitem[{Kidger et~al.(2021{\natexlab{a}})Kidger, Foster, Li, and Lyons}]{kidger2021efficient}
Patrick Kidger, James Foster, Xuechen Li, and Terry~J. Lyons. 2021{\natexlab{a}}.
\newblock \href {https://proceedings.neurips.cc/paper/2021/hash/9ba196c7a6e89eafd0954de80fc1b224-Abstract.html} {Efficient and accurate gradients for neural sdes}.
\newblock In \emph{Advances in Neural Information Processing Systems 34: Annual Conference on Neural Information Processing Systems 2021, NeurIPS 2021, December 6-14, 2021, virtual}, pages 18747--18761.

\bibitem[{Kidger et~al.(2021{\natexlab{b}})Kidger, Foster, Li, and Lyons}]{kidger2021neural}
Patrick Kidger, James Foster, Xuechen Li, and Terry~J. Lyons. 2021{\natexlab{b}}.
\newblock \href {http://proceedings.mlr.press/v139/kidger21b.html} {Neural sdes as infinite-dimensional gans}.
\newblock In \emph{Proceedings of the 38th International Conference on Machine Learning, {ICML} 2021, 18-24 July 2021, Virtual Event}, volume 139 of \emph{Proceedings of Machine Learning Research}, pages 5453--5463. {PMLR}.

\bibitem[{Kidger et~al.(2020{\natexlab{a}})Kidger, Morrill, Foster, and Lyons}]{DBLP:conf/nips/KidgerMFL20}
Patrick Kidger, James Morrill, James Foster, and Terry~J. Lyons. 2020{\natexlab{a}}.
\newblock \href {https://proceedings.neurips.cc/paper/2020/hash/4a5876b450b45371f6cfe5047ac8cd45-Abstract.html} {Neural controlled differential equations for irregular time series}.
\newblock In \emph{Advances in Neural Information Processing Systems 33: Annual Conference on Neural Information Processing Systems 2020, NeurIPS 2020, December 6-12, 2020, virtual}.

\bibitem[{Kidger et~al.(2020{\natexlab{b}})Kidger, Morrill, Foster, and Lyons}]{kidger2020neural}
Patrick Kidger, James Morrill, James Foster, and Terry~J. Lyons. 2020{\natexlab{b}}.
\newblock \href {https://proceedings.neurips.cc/paper/2020/hash/4a5876b450b45371f6cfe5047ac8cd45-Abstract.html} {Neural controlled differential equations for irregular time series}.
\newblock In \emph{Advances in Neural Information Processing Systems 33: Annual Conference on Neural Information Processing Systems 2020, NeurIPS 2020, December 6-12, 2020, virtual}.

\bibitem[{Kossen et~al.(2024)Kossen, Han, Razzak, Schut, Malik, and Gal}]{kossen2024semantic}
Jannik Kossen, Jiatong Han, Muhammed Razzak, Lisa Schut, Shreshth Malik, and Yarin Gal. 2024.
\newblock \href {https://arxiv.org/abs/2406.15927} {Semantic entropy probes: Robust and cheap hallucination detection in llms}.
\newblock \emph{ArXiv preprint}, abs/2406.15927.

\bibitem[{Kwiatkowski et~al.(2019)Kwiatkowski, Palomaki, Redfield, Collins, Parikh, Alberti, Epstein, Polosukhin, Devlin, Lee, Toutanova, Jones, Kelcey, Chang, Dai, Uszkoreit, Le, and Petrov}]{kwiatkowski-etal-2019-natural}
Tom Kwiatkowski, Jennimaria Palomaki, Olivia Redfield, Michael Collins, Ankur Parikh, Chris Alberti, Danielle Epstein, Illia Polosukhin, Jacob Devlin, Kenton Lee, Kristina Toutanova, Llion Jones, Matthew Kelcey, Ming-Wei Chang, Andrew~M. Dai, Jakob Uszkoreit, Quoc Le, and Slav Petrov. 2019.
\newblock \href {https://doi.org/10.1162/tacl_a_00276} {Natural questions: A benchmark for question answering research}.
\newblock \emph{Transactions of the Association for Computational Linguistics}, 7:452--466.

\bibitem[{Levenshtein(1966)}]{levenshtein1966binary}
V~Levenshtein. 1966.
\newblock Binary codes capable of correcting deletions, insertions, and reversals.
\newblock \emph{Proceedings of the Soviet physics doklady}.

\bibitem[{Levinstein and Herrmann(2024)}]{levinstein2024still}
Benjamin~A Levinstein and Daniel~A Herrmann. 2024.
\newblock \href {https://arxiv.org/pdf/2307.00175} {Still no lie detector for language models: Probing empirical and conceptual roadblocks}.
\newblock \emph{Philosophical Studies}, pages 1--27.

\bibitem[{Li et~al.(2022{\natexlab{a}})Li, Du, Zhou, Jing, Zhou, Zeng, Xiao, Zhu, Liu, and Zhang}]{li-etal-2022-ode}
Bei Li, Quan Du, Tao Zhou, Yi~Jing, Shuhan Zhou, Xin Zeng, Tong Xiao, JingBo Zhu, Xuebo Liu, and Min Zhang. 2022{\natexlab{a}}.
\newblock \href {https://doi.org/10.18653/v1/2022.acl-long.571} {{ODE} transformer: An ordinary differential equation-inspired model for sequence generation}.
\newblock In \emph{Proceedings of the 60th Annual Meeting of the Association for Computational Linguistics (Volume 1: Long Papers)}, pages 8335--8351, Dublin, Ireland. Association for Computational Linguistics.

\bibitem[{Li et~al.(2022{\natexlab{b}})Li, Du, Zhou, Jing, Zhou, Zeng, Xiao, Zhu, Liu, and Zhang}]{DBLP:conf/acl/LiDZJZZXZ0Z22}
Bei Li, Quan Du, Tao Zhou, Yi~Jing, Shuhan Zhou, Xin Zeng, Tong Xiao, JingBo Zhu, Xuebo Liu, and Min Zhang. 2022{\natexlab{b}}.
\newblock \href {https://doi.org/10.18653/v1/2022.acl-long.571} {{ODE} transformer: An ordinary differential equation-inspired model for sequence generation}.
\newblock In \emph{Proceedings of the 60th Annual Meeting of the Association for Computational Linguistics (Volume 1: Long Papers)}, pages 8335--8351, Dublin, Ireland. Association for Computational Linguistics.

\bibitem[{Li et~al.(2023{\natexlab{a}})Li, Cheng, Zhao, Nie, and Wen}]{li2023halueval}
Junyi Li, Xiaoxue Cheng, Xin Zhao, Jian-Yun Nie, and Ji-Rong Wen. 2023{\natexlab{a}}.
\newblock \href {https://doi.org/10.18653/v1/2023.emnlp-main.397} {{H}alu{E}val: A large-scale hallucination evaluation benchmark for large language models}.
\newblock In \emph{Proceedings of the 2023 Conference on Empirical Methods in Natural Language Processing}, pages 6449--6464, Singapore. Association for Computational Linguistics.

\bibitem[{Li et~al.(2023{\natexlab{b}})Li, Cheng, Zhao, Nie, and Wen}]{li-etal-2023-halueval}
Junyi Li, Xiaoxue Cheng, Xin Zhao, Jian-Yun Nie, and Ji-Rong Wen. 2023{\natexlab{b}}.
\newblock \href {https://doi.org/10.18653/v1/2023.emnlp-main.397} {{H}alu{E}val: A large-scale hallucination evaluation benchmark for large language models}.
\newblock In \emph{Proceedings of the 2023 Conference on Empirical Methods in Natural Language Processing}, pages 6449--6464, Singapore. Association for Computational Linguistics.

\bibitem[{Li et~al.(2024)Li, Lyu, Geng, Zhu, Panov, and Karray}]{li2024reference}
Qing Li, Chenyang Lyu, Jiahui Geng, Derui Zhu, Maxim Panov, and Fakhri Karray. 2024.
\newblock \href {https://arxiv.org/abs/2408.05767} {Reference-free hallucination detection for large vision-language models}.
\newblock \emph{ArXiv preprint}, abs/2408.05767.

\bibitem[{Li et~al.(2020)Li, Wong, Chen, and Duvenaud}]{DBLP:conf/aistats/LiWCD20}
Xuechen Li, Ting{-}Kam~Leonard Wong, Ricky T.~Q. Chen, and David Duvenaud. 2020.
\newblock \href {http://proceedings.mlr.press/v108/li20i.html} {Scalable gradients for stochastic differential equations}.
\newblock In \emph{The 23rd International Conference on Artificial Intelligence and Statistics, {AISTATS} 2020, 26-28 August 2020, Online [Palermo, Sicily, Italy]}, volume 108 of \emph{Proceedings of Machine Learning Research}, pages 3870--3882. {PMLR}.

\bibitem[{Liang et~al.(2021)Liang, Ouyang, Yan, Wang, Tong, and Zimmermann}]{DBLP:conf/ijcai/LiangOYWTZ21}
Yuxuan Liang, Kun Ouyang, Hanshu Yan, Yiwei Wang, Zekun Tong, and Roger Zimmermann. 2021.
\newblock \href {https://doi.org/10.24963/IJCAI.2021/207} {Modeling trajectories with neural ordinary differential equations}.
\newblock In \emph{Proceedings of the Thirtieth International Joint Conference on Artificial Intelligence, {IJCAI} 2021, Virtual Event / Montreal, Canada, 19-27 August 2021}, pages 1498--1504. ijcai.org.

\bibitem[{Lin et~al.(2022)Lin, Hilton, and Evans}]{lin-etal-2022-truthfulqa}
Stephanie Lin, Jacob Hilton, and Owain Evans. 2022.
\newblock \href {https://doi.org/10.18653/v1/2022.acl-long.229} {{T}ruthful{QA}: Measuring how models mimic human falsehoods}.
\newblock In \emph{Proceedings of the 60th Annual Meeting of the Association for Computational Linguistics (Volume 1: Long Papers)}, pages 3214--3252, Dublin, Ireland. Association for Computational Linguistics.

\bibitem[{Lu et~al.(2019)Lu, Li, He, Sun, Dong, Qin, Wang, and Liu}]{lu2019understanding}
Yiping Lu, Zhuohan Li, Di~He, Zhiqing Sun, Bin Dong, Tao Qin, Liwei Wang, and Tie-Yan Liu. 2019.
\newblock \href {https://arxiv.org/abs/1906.02762} {Understanding and improving transformer from a multi-particle dynamic system point of view}.
\newblock \emph{ArXiv preprint}, abs/1906.02762.

\bibitem[{Lu et~al.(2018)Lu, Zhong, Li, and Dong}]{lu2018beyond}
Yiping Lu, Aoxiao Zhong, Quanzheng Li, and Bin Dong. 2018.
\newblock \href {http://proceedings.mlr.press/v80/lu18d.html} {Beyond finite layer neural networks: Bridging deep architectures and numerical differential equations}.
\newblock In \emph{Proceedings of the 35th International Conference on Machine Learning, {ICML} 2018, Stockholmsm{\"{a}}ssan, Stockholm, Sweden, July 10-15, 2018}, volume~80 of \emph{Proceedings of Machine Learning Research}, pages 3282--3291. {PMLR}.

\bibitem[{Manakul et~al.(2023)Manakul, Liusie, and Gales}]{manakul2023selfcheckgpt}
Potsawee Manakul, Adian Liusie, and Mark Gales. 2023.
\newblock \href {https://doi.org/10.18653/v1/2023.emnlp-main.557} {{S}elf{C}heck{GPT}: Zero-resource black-box hallucination detection for generative large language models}.
\newblock In \emph{Proceedings of the 2023 Conference on Empirical Methods in Natural Language Processing}, pages 9004--9017, Singapore. Association for Computational Linguistics.

\bibitem[{Min et~al.(2023)Min, Krishna, Lyu, Lewis, Yih, Koh, Iyyer, Zettlemoyer, and Hajishirzi}]{MinKLLYKIZH23}
Sewon Min, Kalpesh Krishna, Xinxi Lyu, Mike Lewis, Wen-tau Yih, Pang Koh, Mohit Iyyer, Luke Zettlemoyer, and Hannaneh Hajishirzi. 2023.
\newblock \href {https://doi.org/10.18653/v1/2023.emnlp-main.741} {{FA}ct{S}core: Fine-grained atomic evaluation of factual precision in long form text generation}.
\newblock In \emph{Proceedings of the 2023 Conference on Empirical Methods in Natural Language Processing}, pages 12076--12100, Singapore. Association for Computational Linguistics.

\bibitem[{M{\"u}ndler et~al.(2023)M{\"u}ndler, He, Jenko, and Vechev}]{mundler2023self}
Niels M{\"u}ndler, Jingxuan He, Slobodan Jenko, and Martin Vechev. 2023.
\newblock \href {https://arxiv.org/abs/2305.15852} {Self-contradictory hallucinations of large language models: Evaluation, detection and mitigation}.
\newblock \emph{ArXiv preprint}, abs/2305.15852.

\bibitem[{Oh et~al.(2024)Oh, Lim, and Kim}]{DBLP:journals/corr/abs-2402-14989}
YongKyung Oh, Dongyoung Lim, and Sungil Kim. 2024.
\newblock \href {https://arxiv.org/abs/2402.14989} {Stable neural stochastic differential equations in analyzing irregular time series data}.
\newblock \emph{ArXiv preprint}, abs/2402.14989.

\bibitem[{Orgad et~al.(2025)Orgad, Toker, Gekhman, Reichart, Szpektor, Kotek, and Belinkov}]{orgad2025llms}
Hadas Orgad, Michael Toker, Zorik Gekhman, Roi Reichart, Idan Szpektor, Hadas Kotek, and Yonatan Belinkov. 2025.
\newblock \href {https://openreview.net/forum?id=KRnsX5Em3W} {{LLM}s know more than they show: On the intrinsic representation of {LLM} hallucinations}.
\newblock In \emph{The Thirteenth International Conference on Learning Representations}.

\bibitem[{Pontryagin et~al.(1962)Pontryagin, Mishchenko, Boltyanskii, and Gamkrelidze}]{pontryagin2018mathematical}
Lev~Semenovich Pontryagin, EF~Mishchenko, VG~Boltyanskii, and RV~Gamkrelidze. 1962.
\newblock \emph{The mathematical theory of optimal processes}.
\newblock Routledge.

\bibitem[{Runge(1895)}]{runge1895numerische}
Carl Runge. 1895.
\newblock {\"U}ber die numerische aufl{\"o}sung von differentialgleichungen.
\newblock \emph{Mathematische Annalen}, 46(2):167--178.

\bibitem[{Song et~al.(2024)Song, Xie, Song, Zhu, Huang, Juefei-Xu, and Ma}]{song2024luna}
Da~Song, Xuan Xie, Jiayang Song, Derui Zhu, Yuheng Huang, Felix Juefei-Xu, and Lei Ma. 2024.
\newblock \href {https://arxiv.org/pdf/2310.14211} {Luna: {A} model-based universal analysis framework for large language models}.
\newblock \emph{IEEE Transactions on Software Engineering}.

\bibitem[{Su et~al.(2024)Su, Wang, Ai, Hu, Wu, Zhou, and Liu}]{su-etal-2024-unsupervised}
Weihang Su, Changyue Wang, Qingyao Ai, Yiran Hu, Zhijing Wu, Yujia Zhou, and Yiqun Liu. 2024.
\newblock \href {https://doi.org/10.18653/v1/2024.findings-acl.854} {Unsupervised real-time hallucination detection based on the internal states of large language models}.
\newblock In \emph{Findings of the Association for Computational Linguistics: ACL 2024}, pages 14379--14391, Bangkok, Thailand. Association for Computational Linguistics.

\bibitem[{Taori et~al.(2023)Taori, Gulrajani, Zhang, Dubois, Li, Guestrin, Liang, and Hashimoto}]{alpaca}
Rohan Taori, Ishaan Gulrajani, Tianyi Zhang, Yann Dubois, Xuechen Li, Carlos Guestrin, Percy Liang, and Tatsunori~B. Hashimoto. 2023.
\newblock Stanford alpaca: An instruction-following llama model.
\newblock \url{https://github.com/tatsu-lab/stanford_alpaca}.

\bibitem[{Team et~al.(2024)Team, Mesnard, Hardin, Dadashi, Bhupatiraju, Pathak, Sifre, Rivi{\`e}re, Kale, Love et~al.}]{team2024gemma}
Gemma Team, Thomas Mesnard, Cassidy Hardin, Robert Dadashi, Surya Bhupatiraju, Shreya Pathak, Laurent Sifre, Morgane Rivi{\`e}re, Mihir~Sanjay Kale, Juliette Love, et~al. 2024.
\newblock \href {https://arxiv.org/pdf/2403.08295} {Gemma: Open models based on gemini research and technology}.
\newblock \emph{arXiv preprint arXiv:2403.08295}.

\bibitem[{Touvron et~al.(2023)Touvron, Martin, Stone, Albert, Almahairi, Babaei, Bashlykov, Batra, Bhargava, Bhosale et~al.}]{touvron2023llama}
Hugo Touvron, Louis Martin, Kevin Stone, Peter Albert, Amjad Almahairi, Yasmine Babaei, Nikolay Bashlykov, Soumya Batra, Prajjwal Bhargava, Shruti Bhosale, et~al. 2023.
\newblock \href {https://arxiv.org/abs/2307.09288} {Llama 2: Open foundation and fine-tuned chat models}.
\newblock \emph{ArXiv preprint}, abs/2307.09288.

\bibitem[{Wang et~al.(2023)Wang, Reddy, Mujahid, Arora, Rubashevskii, Geng, Afzal, Pan, Borenstein, Pillai et~al.}]{wang2023factcheck}
Yuxia Wang, Revanth~Gangi Reddy, Zain~Muhammad Mujahid, Arnav Arora, Aleksandr Rubashevskii, Jiahui Geng, Osama~Mohammed Afzal, Liangming Pan, Nadav Borenstein, Aditya Pillai, et~al. 2023.
\newblock \href {https://arxiv.org/abs/2311.09000} {Factcheck-gpt: End-to-end fine-grained document-level fact-checking and correction of llm output}.
\newblock \emph{ArXiv preprint}, abs/2311.09000.

\bibitem[{Wei et~al.(2024)Wei, Yang, Song, Lu, Hu, Tran, Peng, Liu, Huang, Du, and Le}]{safe}
Jerry Wei, Chengrun Yang, Xinying Song, Yifeng Lu, Nathan Hu, Dustin Tran, Daiyi Peng, Ruibo Liu, Da~Huang, Cosmo Du, and Quoc~V. Le. 2024.
\newblock \href {https://arxiv.org/abs/2403.18802} {Long-form factuality in large language models}.
\newblock \emph{ArXiv preprint}, abs/2403.18802.

\bibitem[{Zhang et~al.(2023{\natexlab{a}})Zhang, Qiu, Guo, Deng, Zhang, Zhang, Zhou, Wang, and Fu}]{zhang-etal-2023-enhancing-uncertainty}
Tianhang Zhang, Lin Qiu, Qipeng Guo, Cheng Deng, Yue Zhang, Zheng Zhang, Chenghu Zhou, Xinbing Wang, and Luoyi Fu. 2023{\natexlab{a}}.
\newblock \href {https://doi.org/10.18653/v1/2023.emnlp-main.58} {Enhancing uncertainty-based hallucination detection with stronger focus}.
\newblock In \emph{Proceedings of the 2023 Conference on Empirical Methods in Natural Language Processing}, pages 915--932, Singapore. Association for Computational Linguistics.

\bibitem[{Zhang et~al.(2023{\natexlab{b}})Zhang, Li, Cui, Cai, Liu, Fu, Huang, Zhao, Zhang, Chen et~al.}]{zhang2023siren}
Yue Zhang, Yafu Li, Leyang Cui, Deng Cai, Lemao Liu, Tingchen Fu, Xinting Huang, Enbo Zhao, Yu~Zhang, Yulong Chen, et~al. 2023{\natexlab{b}}.
\newblock \href {https://arxiv.org/abs/2309.01219} {Siren's song in the ai ocean: a survey on hallucination in large language models}.
\newblock \emph{ArXiv preprint}, abs/2309.01219.

\bibitem[{Zhu et~al.(2024)Zhu, Chen, Li, Chen, Ma, Grossklags, and Fritz}]{zhu-etal-2024-pollmgraph}
Derui Zhu, Dingfan Chen, Qing Li, Zongxiong Chen, Lei Ma, Jens Grossklags, and Mario Fritz. 2024.
\newblock \href {https://aclanthology.org/2024.findings-naacl.294} {{P}o{LLM}graph: Unraveling hallucinations in large language models via state transition dynamics}.
\newblock In \emph{Findings of the Association for Computational Linguistics: NAACL 2024}, pages 4737--4751, Mexico City, Mexico. Association for Computational Linguistics.

\end{thebibliography}

\appendix


\section{Fourth-order Runge-Kutta (RK4)}
\label{sec:appendix:rk4}
We can also define a fourth-order Runge-Kutta (RK4) block to be:
\begin{equation}
\small
\begin{split}
z_{t+1} &= z_t + \frac{\Delta t}{6}(Z_1 + 2Z_2 + 2Z_3 + Z_4) \\
Z_1 &= f(t, z_t; \theta_f) \\
Z_2 &= f(t+\frac{\Delta t}{2}, z_t + \frac{\Delta t}{2} Z_1; \theta_f) \\
Z_3 &= f(t+\frac{\Delta t}{2}, z_t + \frac{\Delta t}{2} Z_2; \theta_f) \\
Z_4 &= f(t+\Delta t, z_t + \Delta t Z_3; \theta_f)
\end{split}
\end{equation}

\section{Question Answering Datasets}
\label{sec:app_datasets}
\textbf{TruthfulQA} consists of 873 questions, each with multiple correct and incorrect answers.  For \textbf{HaluEval}, our experiments focused on the ‘QA’ subset comprising 10k records, where each record includes a question accompanied
by both a truthful and a hallucinatory answer. The validation set of \textbf{NQ} consists of 3,610 QA pairs, while the validation set of \textbf{TriviaQA} (rc.nocontext subset) contains 9,960 deduplicated QA pairs. Unlike True-False, we use the Levenshtein~\cite{levenshtein1966binary} distance to select the pair of correct and incorrect answers with the greatest textual difference. These pairs, along with the original questions, form the data used for our experiments. Finally, we generate 1,000 samples in each of the four aforementioned datasets.

\section{Baseline Methods}
\label{sec:app_baseline}
We collect logit-based and consistency-based methods proposed in~\cite{manakul2023selfcheckgpt} to test the effectiveness of models.


\subsection{Logit-based methods}
\label{sec:appendix:logit-based-methods}

To aggregate the uncertainty information obtained at the token level, we employ four metrics to aggregate token-level uncertainty into sentence level. In particular, a sentence-level uncertainty score can be obtained by taking either the maximum or average of the negative loglikelihood $-\log p_{ij}$ in a sentence:
\begin{align}
    \text{MaxProb}(i) &= \max_{j} (-\log p_{ij}),
    \\
    \text{AvgProb}(i) &= -\frac{1}{J_i} \sum_{j = 1}^{J_i} \log p_{ij},
\end{align}
where $p_{ij}$ is the probability of a token at a position $j$ in the sentence $i$ and $J_i$ is the total number of tokens in the considered sentence. Additionally, one can also replace the negative loglikelihood $-\log p_{ij}$ with the entropy  $\mathcal{H}_{ij}$:
\begin{align}
    \text{MaxEnt}(i) &= \max_{j} \mathcal{H}_{ij},
    \\
    \text{AvgEnt}(i) &= \frac{1}{J_i} \sum_{j = 1}^{J_i} \mathcal{H}_{ij},
\end{align}
where $H_{ij}$ is the entropy of the token distribution for the $j$-th token in the sentence $i$.

\subsection{Consistency-based Methods}
\label{sec:appendix:consistency-based-methods}

\noindent\textbf{Unigram.} The concept behind Unigram is to develop a new model that approximates the LVLMs by samples $\{S^1, \dots, S^N\}$ and get the LVLM's token probabilities using this model. As $N$ increases, the new model gets closer to LVLMs. Due to time and cost constraints, we just train a simple $n$-gram model using the samples $\{S^1, \dots, S^N\}$ as well as the main response $R$. We then compare the average and maximum of the negative probabilities of the sentence in response $R$ using the following equations:
\begin{align}
  \mathcal{S}^{\text{Avg}}_{\text{n-gram}}(i) &= -\frac{1}{J_i} \sum_{j = 1}^{J_i} \log \hat{p}_{ij},
  \\
  \mathcal{S}^{\text{Max}}_{\text{n-gram}}(i) &= \max_{j}(-\log \hat{p}_{ij}),
\end{align}
where $\hat{p}_{ij}$ is the probability of a token at position $j$ of a sentence $i$.

\noindent\textbf{Natural Language Inference (NLI)} determines whether a hypothesis follows a premise, classified into either entailment/neutral/contradiction. In this work,  we use DeBERTa-v3-large~\cite{DBLP:conf/iclr/HeGC23} fine-tuned to MNLI as the NLI model. The input for NLI classifiers is typically the premise concatenated to the hypothesis, which for NLI  is the sampled passage $S^n$ concatenated to the sentence to be assessed $r_i$ in the response $R$.
Only the logits associated with the ‘entailment’ and ‘contradiction’ classes are considered,
\begin{equation*}
  P(contradict \mid r_i, S^n) = \frac{\exp\bigl(z_e^{i,n}\bigr)}{\exp\bigl(z_e^{i,n}\bigr) + \exp\bigl(z_c^{i,n}\bigr)},
\end{equation*}
where $z_e^{i,n} = z_e(r_i, S^n)$ and $z_c^{i,n} = z_c(r_i, S^n)$ are the logits of the ‘entailment’ and ‘contradiction’ classes. NLI score for sentence $r_i$ on samples $\{S^1, \dots, S^N\}$ is then defined as,
\begin{equation}
  \mathcal{S}_{\text{NLI}}(i) = \frac{1}{N} \sum_{n=1}^{N}P(contradict \mid r_i, S^n).
\end{equation}

\begin{figure*}[hbt!]
    \centering
    \begin{subfigure}[b]{0.31\linewidth}
    \includegraphics[width=\textwidth]{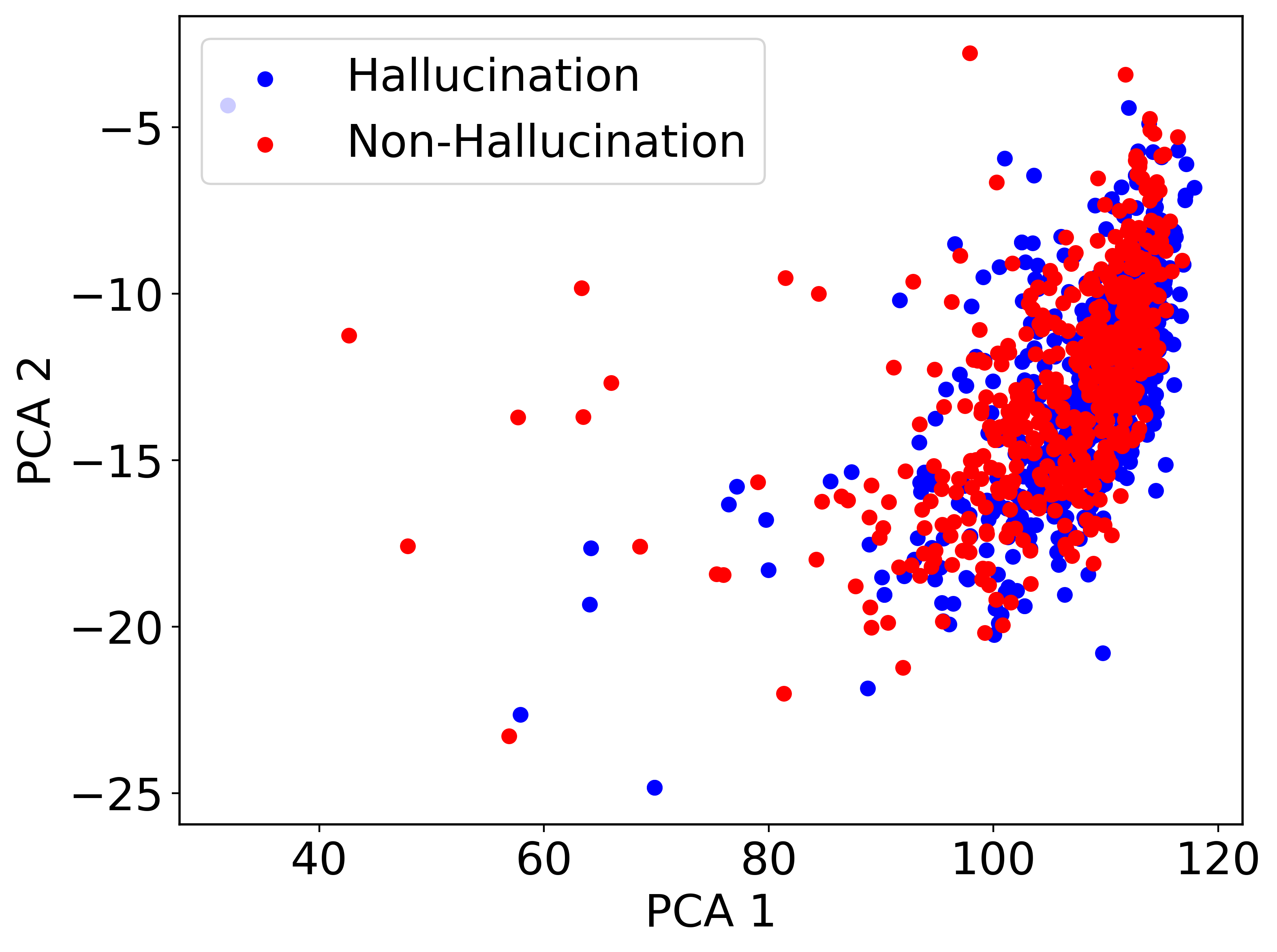}
    \caption{$\text{Company}^{*}$}
    \label{fig:result:logp}
    \end{subfigure}
    \hspace{.00002in}
    \begin{subfigure}[b]{0.31\linewidth}
        \includegraphics[width=\textwidth]{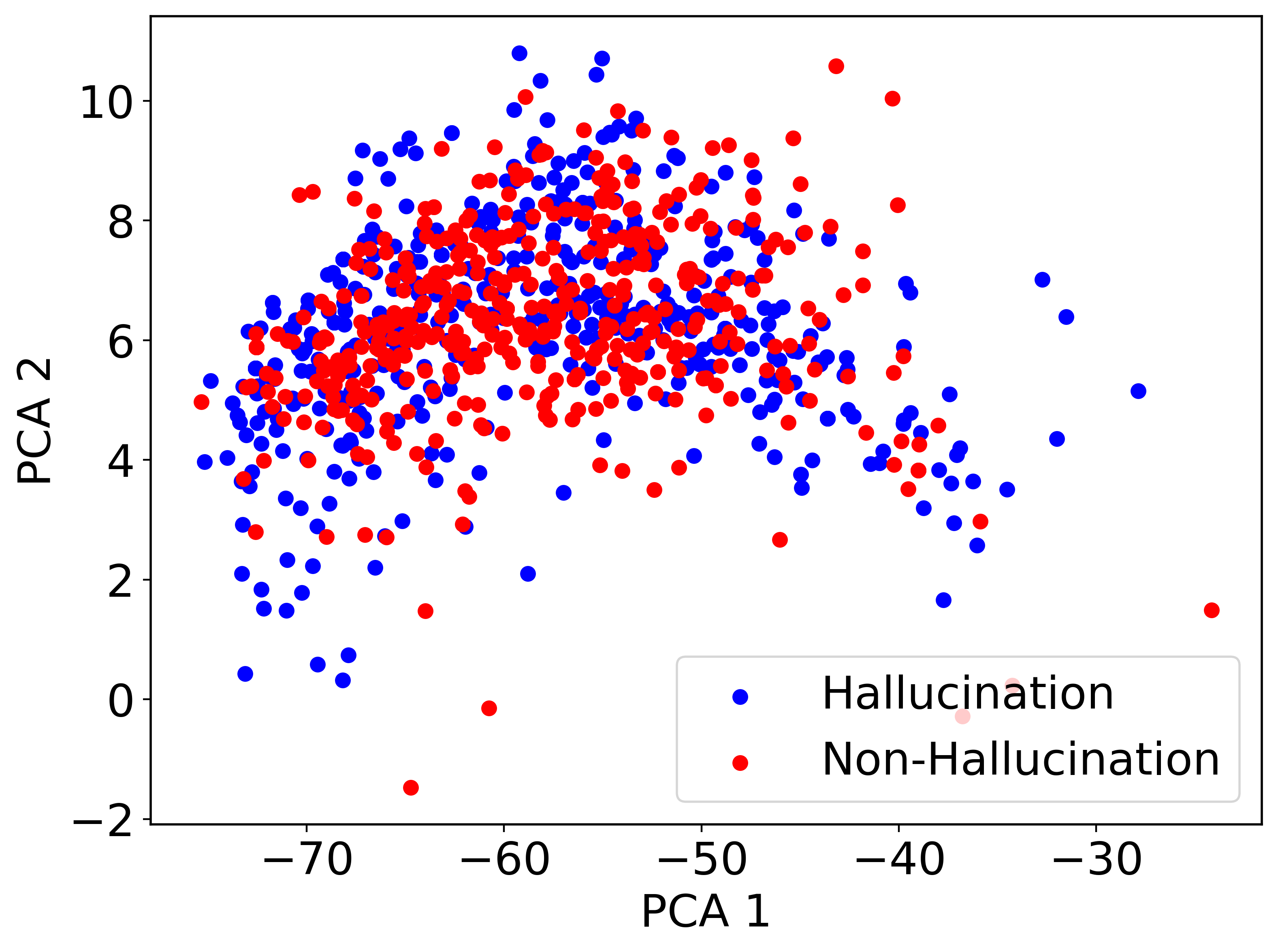}
    \caption{$\text{City}^{*}$}
    \label{fig:result:nli}
    \end{subfigure}
    \hspace{.00002in}
    \begin{subfigure}[b]{0.31\linewidth}
        \includegraphics[width=\textwidth]{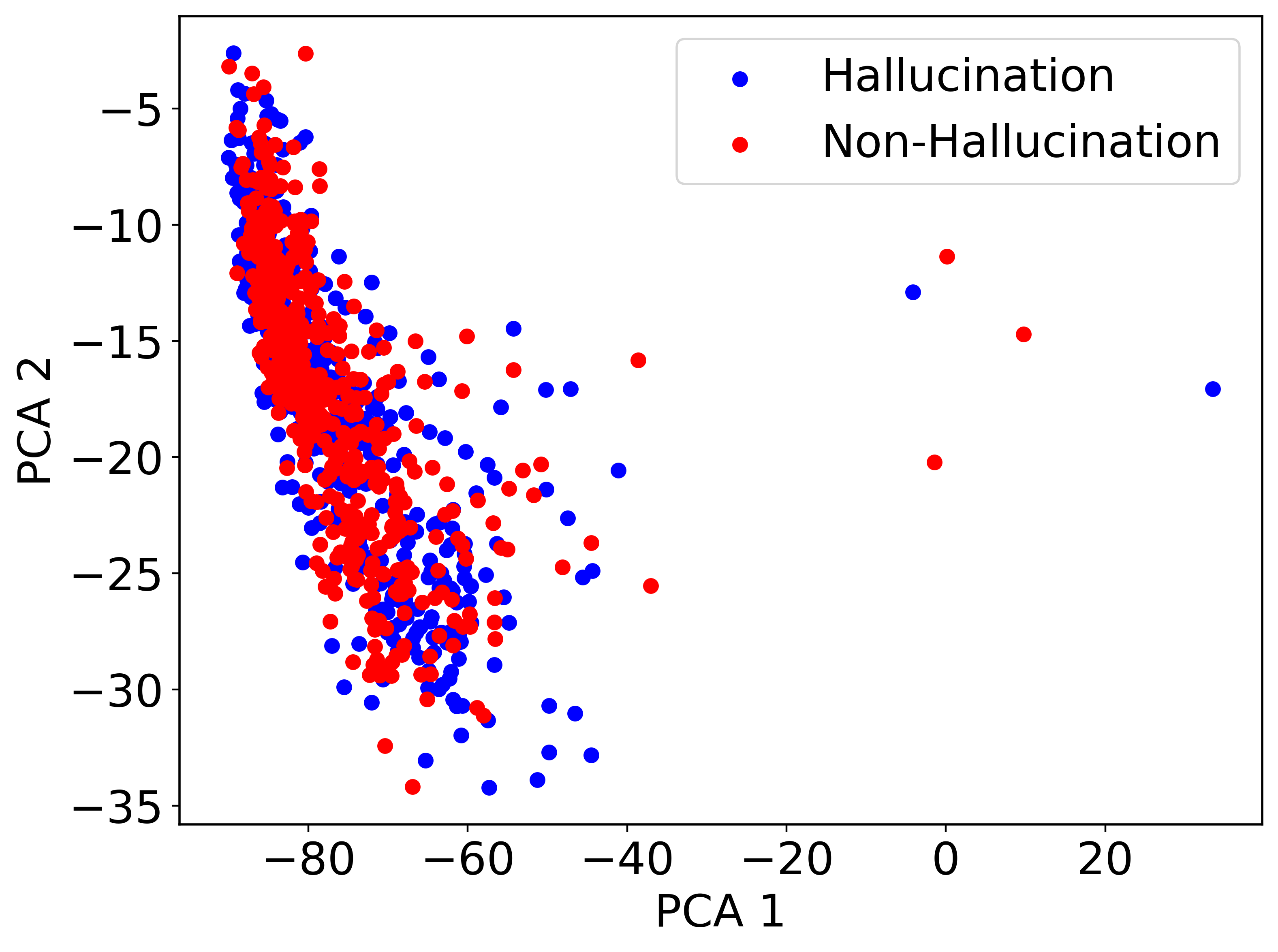}
    \caption{$\text{Invention}^{*}$}
    \label{fig:result:nli}
    \end{subfigure}
    \caption{2D PCA projection of the last hidden layer's embedding for the final token on $\text{Company}^{*}$, $\text{City}^{*}$, $\text{Invention}^{*}$. Blue and red dots represent hallucinations and non-hallucinations, respectively.}
    \label{fig:app:hidden_information:tr_false}
\end{figure*}

\section{Implementation Details}
\label{sec:appendix_implements}

\paragraph{SAPLMA.} We follow the majority of the experimental setup for SAPLMA as described in \cite{azaria-mitchell-2023-internal}. Its classifier employs a feedforward neural network featuring three hidden layers with decreasing numbers of hidden units (1024, 512, 256), all utilizing ReLU activations. The final layer is a sigmoid output. We use the Adam optimizer. The classifier is trained for 20 epochs with a learning rate of 5e-4 and a training batch size of 32. We use about three-quarters of the dataset to train a classifier based on a specific model, and then test its accuracy on the remaining quarter of the same dataset. The training and testing datasets are randomly split.

\paragraph{MIND.} We follow the majority of the experimental setup for MIND as described in~\citet{su-etal-2024-unsupervised}. The MIND classifier utilizes a 4-layer Multilayer Perceptron (MLP) network with a 20\% dropout applied to the initial layer. The network architecture features decreasing hidden layer sizes of 256, 128, 64, and 2 for each layer. The Rectified Linear Unit (ReLU) activation function is used, with a learning rate of 5e-4, a weight decay of 1e-5, and a training batch size of 32.

\paragraph{Probe@Exact.} We follow the majority of the experimental setup for Probe@Exact as described in~\citet{orgad2025llms}.  We employ the logistic regression model from the scikit-learn library as the probing classifier. For Question Answering Datasets, we use the same method as \citet{orgad2025llms} to detect and utilize exact answer tokens. However, for True-False Datasets, we select key tokens as the exact answer tokens.

\paragraph{P(True).} The prompt that we use for \textit{P(True)}~\citet{kadavath2022language} is as follows:

\begin{tcolorbox} 
Given the following question and answer, your objective is to determine if the answer correctly answers the question. You should give the probability that your think answer is correct.

Question: [\textbf{Question}]

Answer: [\textbf{[Answer]}]
\end{tcolorbox}

\paragraph{LLM Configuration.} For the selected LLMs, we download the model parameters directly from their official Hugging Face repositories. The generation process follows each model's official default configurations.

\section{2D PCA Projection on Other Datasets}
\label{sec:appendix_more_dataset}

Figure~\ref{fig:app:hidden_information:tr_false} shows the 2D PCA projection of the last hidden layer’s embedding for the final token on $\text{Company}^{*}$, $\text{City}^{*}$, $\text{Invention}^{*}$. It reveals a significant overlap between correct and incorrect statements.

\end{document}